\documentclass[11pt]{article}

\usepackage{acl}
\usepackage{times}
\usepackage{latexsym}
\usepackage{booktabs}
\usepackage{multirow}

\usepackage[T1]{fontenc}
\usepackage[T5]{fontenc}

\usepackage[utf8]{inputenc}
\usepackage{amsmath}
\usepackage{microtype}

\usepackage{inconsolata}
\usepackage{amsmath}
\usepackage{cleveref}
\usepackage{graphicx}

\usepackage{setspace}

\graphicspath{{img/}}

\title{When Cultures Meet: Multicultural Text-to-Image Generation}

 \author{Parth Bhalerao \hspace{5pt} 
 Oana Ignat \hspace{5pt}
 Mounika Yalamarty  \hspace{5pt}
Brian Trinh \\
Santa Clara University - Santa Clara, USA  \\
 \textit{\{pbhalerao, oignat\}@scu.edu} \\  }

\begin{document}
\maketitle
\begin{abstract}
Text-to-image generation models have achieved strong performance in culturally homogeneous settings, yet their ability to generate \emph{multicultural scenes}—where people and landmarks originate from different cultures—remains largely unexplored.
We introduce \emph{multicultural text-to-image generation} as a new task and present the first benchmark designed to study this setting. Our dataset contains 9,000 images spanning five countries, three age groups, two genders, 25 historical landmarks, and five languages. Using this benchmark, we analyze the behavior of state-of-the-art text-to-image models across multiple dimensions, including alignment, image quality, aesthetics, knowledge, and fairness.
As one strategy for composing cultural and demographic information, we explore \texttt{MosAIG}, a \underline{M}ulti-\underline{A}gent framework that enhances multicultural \underline{I}mage \underline{G}eneration by leveraging LLMs with distinct cultural personas. Our analysis shows that richer prompt composition can improve image quality and cultural grounding compared to simple prompts, while revealing substantial disparities across languages and demographic groups. We release our dataset and code at \url{https://github.com/AIM-SCU/MosAIG}.
\end{abstract}

\section{Introduction}

Societies worldwide are increasingly diverse, shaped by global travel and migration~\cite{migration}. 
This multicultural reality poses important challenges for Artificial Intelligence (AI), where robust representation of diverse populations is essential for equity and inclusivity~\cite{hershcovich-etal-2022-challenges, naous2023having, mihalcea2024ai}. 
However, most datasets used for text-to-image generation focus on narrow demographics—predominantly Western, adult, and male—and largely depict single-culture scenarios (e.g., \textit{a Chinese temple}, \textit{an Indian market})~\cite{liu2024cultural, kannen2024beyond}. 
Such representations fail to capture common multicultural interactions, for example \textit{a Chinese girl visiting the Golden Gate Bridge}, limiting the applicability of text-to-image models in real-world, culturally diverse settings~\cite{hershcovich-etal-2022-challenges, bhatia-etal-2024-local}.

\begin{figure}
\centering
\includegraphics[width=\linewidth]{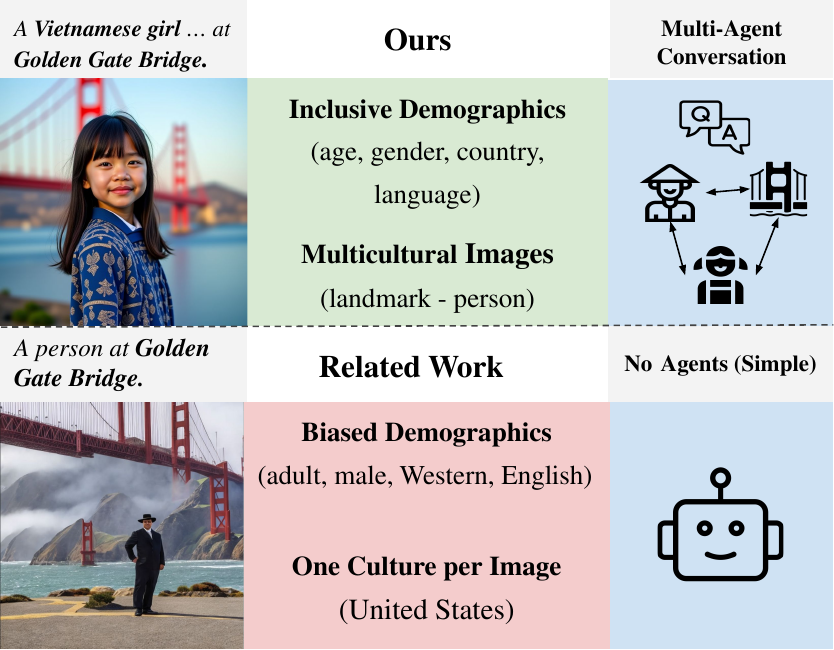}
\caption{Most existing datasets emphasize singular cultural contexts (e.g., the Golden Gate Bridge depicted primarily with American visitors or as a standalone monument). In contrast, real-world scenes often involve people from different cultural backgrounds sharing spaces and experiences. Modeling such multicultural interactions enables richer and more realistic image generation.}
\vspace{-1.2em}
\label{fig:main_idea}
\end{figure}

Despite recent efforts to evaluate cultural awareness in vision--language models, existing benchmarks and analyses primarily consider one culture per image. To date, there has been no systematic study of \emph{multicultural text-to-image generation}, where elements from different cultural origins—such as a person and a landmark—co-exist within the same visual scene. We introduce this setting as a new task and study how current text-to-image models handle cultural composition, demographic variation, and cross-cultural grounding.

Specifically, we examine two key dimensions: (1) the demographic attributes of the depicted person, and (2) the multicultural interaction between the person and the landmark (e.g., the Golden Gate Bridge). We consider four demographic aspects—age, gender, nationality, and language—together with cross-cultural landmarks (\Cref{fig:main_idea}). By systematically exploring these factors and their intersections, we aim to better understand the strengths and limitations of state-of-the-art text-to-image models in multicultural settings.

Our work is guided by the following research questions:
\begin{description}
\item [RQ1:] How accurately do state-of-the-art text-to-image models depict people from one culture within the context of a landmark associated with a different culture?
\vspace{-0.3em}
\item [RQ2:] How does text-to-image generation performance vary across demographic groups and languages?
\vspace{-0.5em}
\item [RQ3:] What modeling strategies help improve multicultural text-to-image generation?
\end{description}

Contributions. \textbf{First, we release the first dataset of 9,000 images designed to study multicultural interactions}, depicting people and landmarks from different cultures across five countries, three age groups, two genders, 25 historical landmarks, and five languages. 
\textbf{Second, we explore \texttt{MosAIG}, a multi-agent prompting framework that decomposes cultural and demographic aspects} during caption construction as an effective strategy for addressing the task. 
\textbf{Finally, we analyze multicultural text-to-image generation through automated and human evaluation, revealing demographic and linguistic disparities.}

\vspace{-0.3em}
\section{Related Work}

\paragraph{Cultural Evaluation in Language and Vision Models.}
Recent work has made substantial progress in modeling and evaluating cultural awareness in language models through large multilingual benchmarks~\cite{pawar2024survey, romanou2024include, singh2024global}. 
In the vision--language domain, benchmarks such as CVQA~\cite{romero2024cvqaculturallydiversemultilingualvisual} and GlobalRG~\cite{bhatia-etal-2024-local} evaluate culturally grounded question answering, retrieval, and visual grounding. 
While multi-agent approaches have been explored for cross-cultural reasoning in multimodal systems~\cite{guo2024large, han2024llm}, prior work such as MosAIC~\cite{bai2024power} focuses on image captioning in single-culture settings, where models describe visual content post hoc. In contrast, we study \emph{text-to-image generation in multicultural settings}, where models must jointly reason about and synthesize multiple cultural, demographic, and landmark-specific cues into a single coherent visual scene. This setting constitutes a particularly challenging test of cultural competence, as failures in grounding, representation, or composition are directly reflected in the generated images.

\paragraph{Text-to-Image Generation Models and Benchmarks.}
Text-to-image generation has advanced rapidly with models such as Stable Diffusion-XL~\cite{podell2023sdxl}, DALL·E~3~\cite{betker2023improving}, and FLUX~\cite{flux2023}. 
Agentic approaches like GenArtist~\cite{wang2024genartist} focus on unified generation and editing pipelines, whereas our work emphasizes multicultural and multilingual evaluation rather than model design.
Existing benchmarks, including TIFA~\cite{hu2023tifa}, GenEval~\cite{ghosh2024geneval}, and GenAIBench~\cite{lin2025evaluating}, primarily assess technical properties such as realism, faithfulness, and compositionality. 
More recent efforts, such as HEIM~\cite{lee2024holistic}, incorporate socially situated dimensions including bias, toxicity, and aesthetics~\cite{hartwig2024evaluating}, but do not explicitly address multicultural scene composition.

\paragraph{Cultural and Linguistic Gaps in Text-to-Image Generation.}
Despite these advances, most text-to-image systems and evaluations remain centered on a limited set of high-resource languages, leaving many linguistic communities underserved. 
While models such as Taiyi-Diffusion-XL~\cite{wu2024taiyi} and AltDiffusion~\cite{ye2024altdiffusion} expand multilingual input coverage, a broader cultural gap persists~\cite{liu2024cultural}, as existing benchmarks rarely capture cross-cultural interactions or multicultural contexts~\cite{hershcovich-etal-2022-challenges, mihalcea2024ai, saha-etal-2025-meta}.

\paragraph{Data Diversity and Cultural Competence.}
Recent work has begun to assess cultural competence in text-to-image generation. For example, CUBE~\cite{kannen2024beyond} and TIFA~\cite{hu2023tifa} evaluate cultural awareness and diversity, but remains limited to single-culture depictions per image. 
To our knowledge, no prior work systematically studies \emph{multicultural image generation}, where multiple cultures co-exist within the same scene. Our work addresses this gap by introducing a benchmark and analysis framework for multicultural text-to-image generation.

\vspace{-0.6em}

\section{Multicultural Image Generation}
\textit{Culture is a multifaceted concept, meaning different things to different people at different times}~\cite{adilazuarda-etal-2024-towards}. 
In this work, we adopt the definition of \citet{nguyen2023extracting} and focus on \emph{visual cultural elements}, such as clothing and historical landmarks.

We introduce \emph{multicultural image generation} as a new task that evaluates how text-to-image models represent elements from multiple cultures within a single image—specifically, a person from one cultural background depicted in the context of a landmark from another. In addition to cultural origin, we examine demographic attributes and their intersections, including age, gender, and language\footnote{All demographics are listed in Appendix \Cref{tab:age_gender_country_landmark}.}.
To address this task, we introduce \texttt{MosAIG}, a \textit{novel framework} for \underline{M}ulti-\underline{A}gent \underline{I}mage \underline{G}eneration, as illustrated in \Cref{fig:pipeline}.
Our framework generates comprehensive image captions that are used to generate more accurate multicultural images using off-the-shelf image generation models.
This framework is built around a multi-agent interaction model, as described below.

\subsection{Multi-Agent Interaction Model}

\begin{figure}
\centering
\includegraphics[width=1\linewidth]{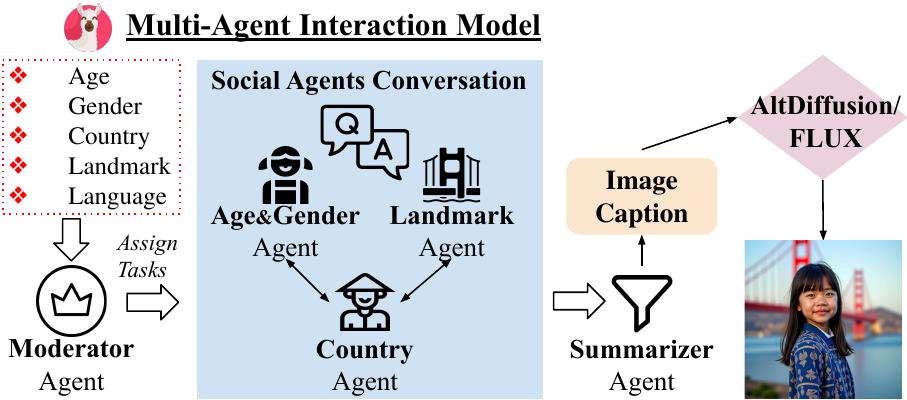}
\caption{Overview of \texttt{MosAIG}, our framework for \underline{M}ulti-\underline{A}gent \underline{I}mage \underline{G}eneration. The framework includes a multi-agent interaction model that generates an image caption from demographic information (person age, gender, country, landmark, and caption language), which is then used by an image generation model to create a multicultural image of a landmark and a person.}
\vspace{-1.2em}
\label{fig:pipeline}
\end{figure}

We introduce a multi-agent setup to emulate collaboration between demographically diverse groups.
Our setup contains five agents, with specific roles: one Moderator Agent, three Social Agents, and one Summarizer Agent, as illustrated in \Cref{fig:pipeline}.

\noindent\textbf{Moderator Agent.} The Moderator Agent obtains demographic (age, gender, nationality) information about the person, the name of the landmark (e.g., Taj Mahal), and the language of the caption as input.
The Moderator Agent then assigns tasks to the Social agents, instructing them to focus on the visually relevant aspects of the input information. 

\noindent\textbf{Social Agents.} The Social Agents interact by asking each other relevant questions to create an image caption according to the information provided by the Moderator Agent. Each Social Agent assumes a \textit{persona}: the first agent represents the culture of the person in the image, the second agent represents the age and gender of the person, and the last agent represents the historical landmark.
Each agent generates an initial description of their persona. Then, by interacting through multiple rounds of question-answering conversations, each agent creates a more comprehensive image description.

\noindent\textbf{Summarizer Agent.} The Summarizer Agent collects the three descriptions from the Social Agents and summarizes them into a final image caption with a maximum length of 77 tokens.

\noindent\textbf{Social Agents Conversation.}
At the start, the three Social Agents—Country Agent, Landmark Agent, and Age-Gender Agent—receive demographic information and tasks from the Moderator Agent.
The Country Agent processes nationality information and describes traditional attire, which is then evaluated by the Age-Gender Agent (e.g., ``Is this attire suitable for a young female?''). Adjustments, such as modifying the color or style of a garment to suit the individual's age, are made accordingly.
The Landmark Agent describes the landmark architecture, and its descriptions are refined based on feedback from the Country Agent (e.g., ``How do Vietnamese visitors typically interact with this landmark?''), ensuring cultural authenticity.
The Age-Gender Agent generates demographic descriptions, which are cross-checked with the Country Agent to ensure culturally appropriate accessories and mannerisms.
After two rounds of conversation, the agents enhance and refine the descriptions with culturally sensitive and contextually rich details. Once the iterative improvement process is complete, the refined descriptions are passed to the Summarizer Agent, which condenses them into a final 77-token prompt capturing the cultural and contextual nuances.
The prompts and implementation details are provided in \Cref{fig:prompts} and \Cref{sec:model}.

\subsection{Image Generation Models}

We evaluate our generated image captions using two different state-of-the-art image generation models: AltDiffusion~\cite{ye2024altdiffusion} and FLUX~\cite{flux2023}.
\vspace{-0.6em}

\paragraph{AltDiffusion.}
AltDiffusion\footnote{\url{https://huggingface.co/BAAI/AltDiffusion-m18}}~\cite{ye2024altdiffusion} is one of the very few multilingual open-source image generation models. The model aligns multilingual language models with diffusion models to generate high-quality images from text across multiple languages. The model builds on CLIP~\cite{radford2021learning}, replacing its text encoder with XLM-R~\cite{conneau2019unsupervised} and employing a two-stage training process that combines teacher learning and contrastive learning.
AltDiffusion supports 18 different languages; we select five—English, German, Hindi, Spanish, and Vietnamese—based on the annotators' expertise. The model processes text inputs with a maximum length of 77 tokens.  
\vspace{-0.6em}
\paragraph{FLUX.}
FLUX.1-dev\footnote{\url{https://huggingface.co/black-forest-labs/FLUX.1-dev}}~\cite{flux2023} is a state-of-the-art, widely used, open-source text-to-image model designed for English-language prompts. Due to computational constraints, we employ Flux.1 Lite\footnote{\url{https://huggingface.co/Freepik/flux.1-lite-8B-alpha}}~\cite{flux1-lite}, an 8B-parameter transformer model, more efficient variant distilled from FLUX.1-dev. 

\vspace{-0.6em}
\subsection{Simple vs. Multi-Agent Image Generation}

Simple models generate images based on predefined captions, whereas multi-agent models utilize dynamically generated captions derived from multi-agent interactions.
For instance, when provided with demographic details such as ``Vietnamese'' (nationality), ``child'' (age), ``female'' (gender), ``Golden Gate Bridge'' (landmark), and ``English'' (caption language), the resulting image captions differ between the two approaches. 
Multi-agent models generate captions that provide richer contextual information, including detailed descriptions of the landmark’s architecture and surroundings, as well as a more nuanced depiction of the person's appearance, particularly focusing on clothing and facial features, as shown below\footnote{All the captions are shown in our code repository.}.

\begin{description}
\small
\item [Simple caption:] {\it A Vietnamese girl wearing traditional attire, standing in front of the Golden Gate Bridge.} 
\vspace{-0.3em}
\item [Multi-agent caption:] {\it A 12-year-old Vietnamese girl in Áo Dài, standing on the Golden Gate Bridge, with the San Francisco Bay's blue waters and the bridge's orange-red towers in the background.} 
\end{description}

\vspace{-0.6em}
\section{Evaluation and Results}

We employ both automated metrics and human evaluation to provide a holistic and comprehensive assessment of the generated images.

\vspace{-0.6em}

\subsection{Evaluation Metrics}

We adopt a set of automated evaluation metrics that assess text-to-image generation along five complementary dimensions: \textbf{Alignment}, \textbf{Quality}, \textbf{Aesthetics}, \textbf{Knowledge}, and \textbf{Fairness}. Together, these metrics capture both technical properties of generation—such as semantic correspondence and visual fidelity—as well as socially situated aspects, including representational consistency across demographic groups~\cite{lee2024holistic}. Given the known limitations of any single automatic metric, we combine multiple evaluators and complement them with human judgment.

\noindent\textbf{Alignment.}
We measure text--image alignment using CLIPScore~\cite{hessel2021clipscore}, a widely adopted, reference-free metric that computes cosine similarity between joint text and image embeddings and enables scalable evaluation. While CLIPScore provides a useful proxy for semantic correspondence, it does not capture all aspects of visual grounding or compositional correctness. Accordingly, we interpret alignment scores cautiously and complement them with human evaluation, and we encourage future work to incorporate image-based classifiers for more direct assessment of visual attribute realization (see Limitations). CLIPScore values range from $-1$ to $+1$, with higher values indicating stronger alignment.

\noindent\textbf{Quality.}
We assess image quality using the Inception Score (IS)~\cite{salimans2016improved}, which evaluates both visual fidelity and output diversity based on predictions from an Inception-v3 classifier. Lower scores typically reflect poor realism or limited variation, while higher scores indicate more realistic and diverse images. Although IS does not directly assess semantic correctness, it provides a complementary signal for overall visual plausibility.

\noindent\textbf{Aesthetics.}
Aesthetic quality captures visual appeal beyond semantic correctness, including sharpness, color harmony, composition, and overall clarity. We use a SigLIP-based aesthetic predictor\footnote{\url{https://github.com/discus0434/aesthetic-predictor-v2-5}}, which assigns scores on a 1--10 scale. This metric prioritizes perceptual attributes and may be less sensitive to semantic or cultural accuracy, making it particularly informative when interpreted alongside alignment and knowledge metrics.

\noindent\textbf{Fairness.}
We evaluate fairness as the consistency of model performance across demographic substitutions. Following prior work, we modify captions by changing demographic attributes such as \textit{gender}, \textit{age}, or \textit{nationality}, while keeping all other elements fixed. Given an original caption--image pair $(c, I)$, we construct a modified caption $c'$ (e.g., replacing ``boy'' with ``girl'') and generate a corresponding image $I'$. Fairness is measured as the absolute difference in alignment:
\[
\Delta S = |S(c, I) - S(c', I')|.
\]
Lower values of $\Delta S$ indicate more consistent behavior across demographic groups, while higher values suggest potential representational disparities. This metric captures relative changes rather than absolute bias and is best interpreted comparatively across models.

\noindent\textbf{Knowledge.}
We assess world knowledge by evaluating a model’s sensitivity to landmark identity. Given a caption--image pair $(c, I)$, we replace the referenced historical landmark in the caption while keeping the image fixed, yielding $(c', I)$. We compute:
\[
\Delta S = S(c, I) - S(c', I).
\]
A model with stronger landmark knowledge should exhibit larger alignment differences when the caption references an incorrect landmark.

\subsection{Multi-Agent Interaction Results}\label{sec:multiagent}

\begin{figure}
\centering
\includegraphics[width=1\linewidth]{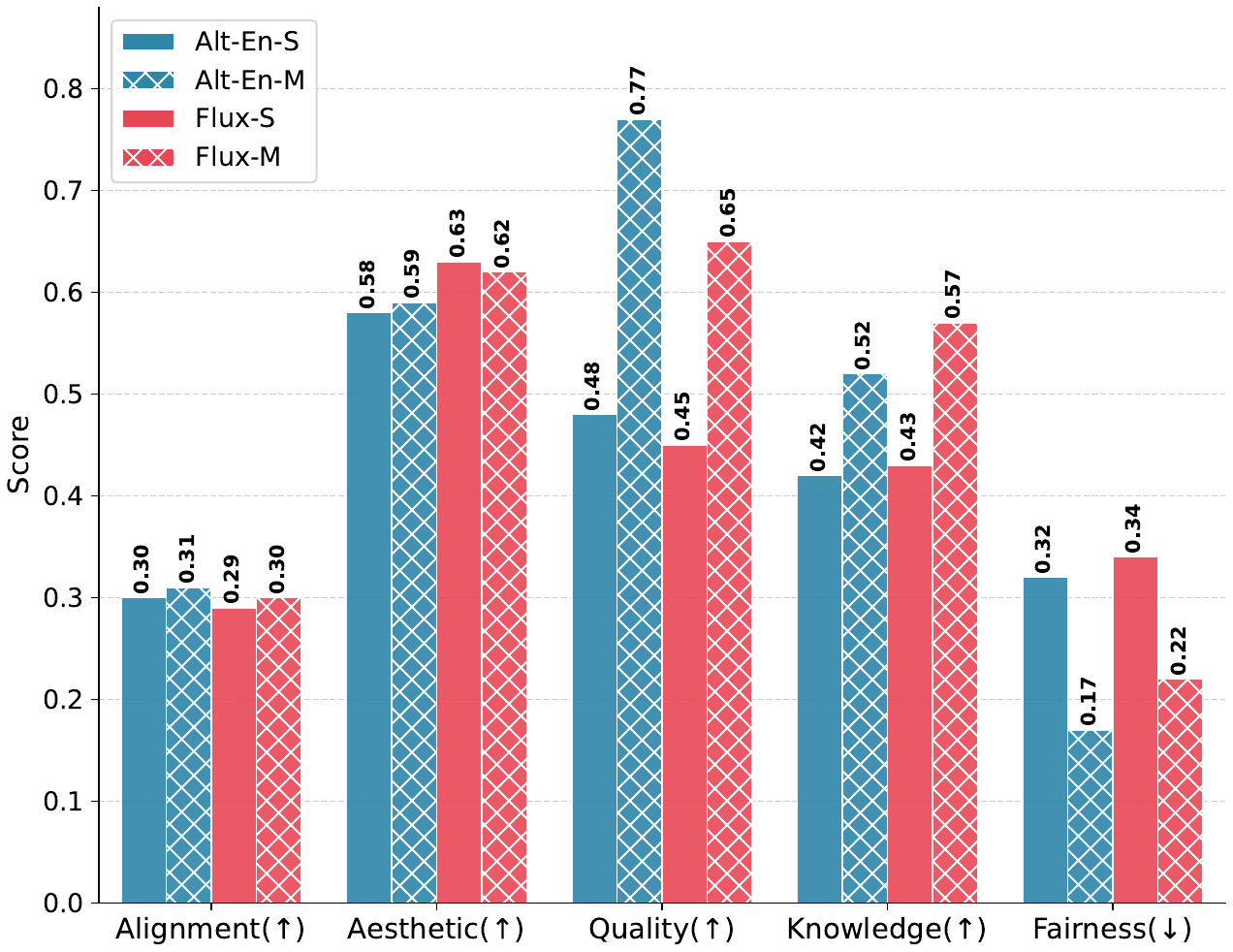}
  \vspace{-1.5em}
\caption{
Multi-agent prompting improves \textit{Quality}, \textit{Knowledge}, and \textit{Fairness} relative to simple prompts, while achieving comparable performance in \textit{Alignment} and \textit{Aesthetics}. Scores are normalized to [0–1]; higher is better except for Fairness.}
\vspace{-1em}
\label{fig:main_plot_bar}
\end{figure}

\Cref{fig:main_plot_bar} compares multi-agent and simple prompting strategies across five evaluation dimensions. Overall, multi-agent prompting yields consistent improvements in \textbf{Quality}, \textbf{Knowledge}, and \textbf{Fairness}, while achieving comparable performance in \textbf{Alignment} and \textbf{Aesthetics}.

The largest gains are observed in \textbf{Quality}. Multi-agent models achieve substantially higher scores than simple prompts (0.77 vs.\ 0.48 for \texttt{Alt-En} and 0.65 vs.\ 0.45 for \texttt{Flux-En}), indicating more visually coherent and photorealistic outputs. We attribute these improvements to richer prompt composition that more explicitly specifies culturally grounded visual details, which appears to reduce under-specified or visually inconsistent generations. Across both prompting strategies, \texttt{Alt} consistently attains higher Quality scores than \texttt{Flux}, likely reflecting differences in background sharpness and image fidelity between the underlying generation models. In contrast, \textbf{Aesthetic} scores remain largely unchanged. This suggests that multi-agent prompting primarily affects semantic and compositional correctness rather than stylistic attributes emphasized by aesthetic predictors, which tend to prioritize surface-level visual appeal.

Multi-agent prompting also improves \textbf{Knowledge} and \textbf{Fairness}. Knowledge scores increase from 0.42 to 0.52 for \texttt{Alt-En} and from 0.43 to 0.57 for \texttt{Flux-En}, indicating stronger sensitivity to landmark-specific information when cultural context is more explicitly specified. Fairness scores—where lower values indicate smaller performance disparities—are substantially reduced (0.17 vs.\ 0.32 for \texttt{Alt-En} and 0.22 vs.\ 0.34 for \texttt{Flux-En}), suggesting more consistent behavior across demographic substitutions. These results indicate that decomposing cultural and demographic cues during prompt construction can mitigate uneven performance across social groups.

Improvements in \textbf{Alignment} are more modest and not statistically significant in aggregate. However, disaggregated analysis reveals consistent gains across several demographic dimensions, including \textit{adults} (0.30 vs.\ 0.27), \textit{females} (0.31 vs.\ 0.28), and multiple countries such as \textit{Germany}, \textit{India}, and \textit{Vietnam} (see \Cref{sec:results}). This pattern suggests that richer cultural specification can improve semantic correspondence for particular population groups, even when overall alignment scores remain similar.

\begin{figure}
\centering
\includegraphics[width=0.9\linewidth]{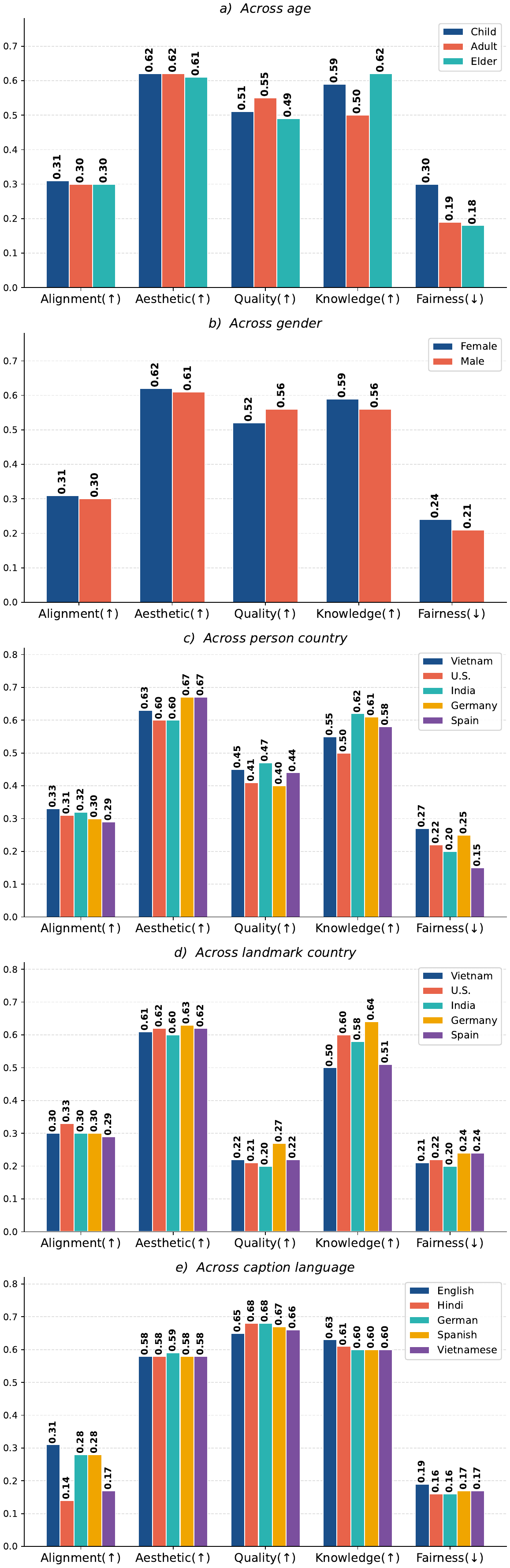}
  \vspace{-1em}
\caption{
Ablation studies on (a) person age, (b) person gender, (c) person country, (d) landmark country, (e) caption language using the best overall model, the \texttt{Multi-agent English Flux-M}  (a-d) and \texttt{Multi-agent Multilingual Alt-M} (e). Performance across all five metrics—Alignment, Aesthetic, Quality, Knowledge, and Fairness—reveals significant variation across these demographic categories.}
\label{fig:ablations}
\end{figure}

\subsection{Ablation Studies}
We also perform ablation studies to
assess \texttt{MosAIG’s} performance across demographics.
\vspace{-0.5em}
\paragraph{a) Person Age.}
\Cref{fig:ablations} a) shows that Image Quality varies by age group, with Adults achieving the highest quality (0.55), followed by Children (0.51) and Elders (0.49).
The model is also fairer when depicting Elders (0.18) and Adults (0.19) compared to Children (0.30).
\vspace{-0.7em}
\paragraph{b) Person Gender.}
\Cref{fig:ablations} b) shows that Knowledge and Image Quality varies by gender, with Males achieving higher quality (0.56) than Females (0.52). However, the model is fairer when depicting Males (0.21) than Females (0.24). The other metrics remain consistent across both groups.
\vspace{-0.7em}
\paragraph{c) Person Country.}
\Cref{fig:ablations} c) shows that model performance varies by person’s country. Alignment is highest for Indian people (0.32) and lowest for Spanish people (0.29). Similarly, Image Quality is highest for Indian people (0.47) and lowest for German people (0.41). The model is also fairest when depicting Spanish (0.15) and least fair for Vietnamese (0.27).
\vspace{-0.7em}
\paragraph{d) Landmark Country.} \Cref{fig:ablations} d) shows that model performance varies by landmark country. The most notable difference is in the Knowledge metric, with German landmarks being the most well-known (0.64), followed by U.S. (0.60), Indian (0.54), Vietnamese (0.50), and Spanish (0.51). Alignment is highest for U.S. landmarks (0.33) and lowest for Spanish landmarks (0.29).
\paragraph{e) Caption Language.}
\vspace{-0.7em}
\Cref{fig:ablations} e) shows that model performance varies by caption language, with English achieving the highest Alignment (0.31) and Knowledge (0.63), while Hindi and Vietnamese score the lowest (0.14 and 0.43, respectively). This disparity may stem from differences in training data availability, as model performance moderately correlates with dataset size (Pearson coefficient: 0.5), estimated from CommonCrawl \cite{wenzek-etal-2020-ccnet}.

\begin{figure}[h]
\centering
\includegraphics[width=1\linewidth]{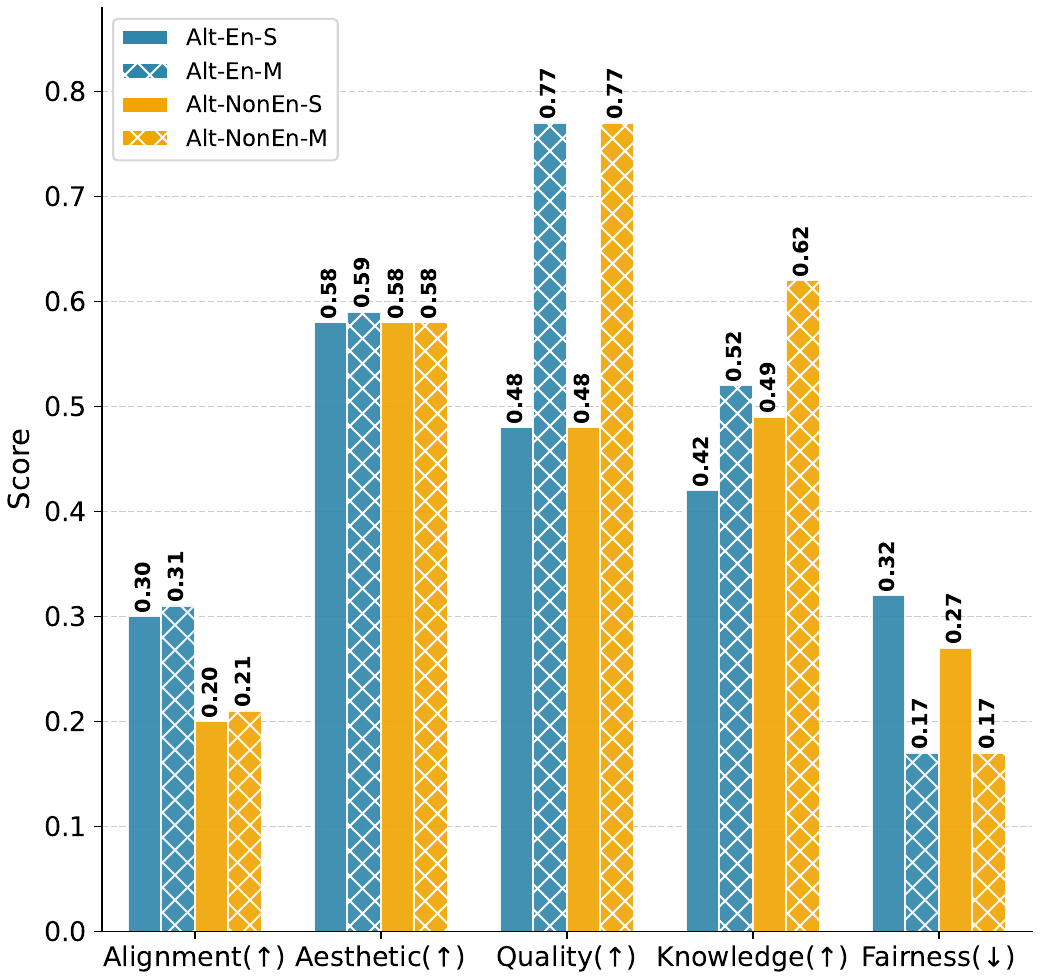}
\caption{
English vs. Multilingual Performance. Models with English captions as input (Alt-En-S, Alt-En-M) achieve higher scores than non-English (Alt-NonEn-S, Alt-NonEn-M) in Alignment (0.30 vs. 0.20), while performing comparably across Aesthetics and Quality metrics. Knowledge and Fairness performance is higher for non-English models}
\label{fig:alt_plot_bar}
\end{figure}

As shown in \Cref{fig:alt_plot_bar}, English models (Alt-En-S, Alt-En-M) achieve substantially higher Alignment (0.30 vs. 0.20) compared to non-English models (Alt-NonEn-S, Alt-NonEn-M), while performing comparably in Aesthetics and Quality. Notably, non-English models achieve higher Knowledge (0.62 vs. 0.52) and lower Fairness scores (0.17 vs. 0.31), indicating more consistent behavior across demographic substitutions, suggesting that multilingual prompts may encode more culturally specific information despite lower overall alignment.

\vspace{-0.7em}
\paragraph{f) Intersectionality.}
Examining a single demographic category, such as race or gender, may overlook nuanced inequalities \cite{field_survey_2021}. To address this, we analyze the intersectionality of age and gender, person and landmark country, and language and person country. We measure Alignment and analyze other metrics across various demographic intersections, as detailed in \Cref{sec:intersection}.

\begin{figure}
\centering
\includegraphics[width=\linewidth]{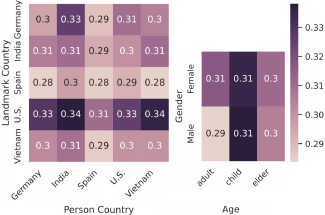}
  \vspace{-0.8em}
\caption{
Alignment with best overall model, \texttt{Flux-M}, over person-landmark (left) and gender-age (right).}
\vspace{-1.2em}
\label{fig:heatmap_person_landmark}
\end{figure}

\noindent\textbf{Age and Gender.}
\Cref{fig:heatmap_person_landmark} (right) shows that Alignment performance varies by gender for generating adult images, with males having a lower score (0.29) compared to females (0.31). The performance for child and elder categories remains consistent across gender.

\noindent\textbf{Person and Landmark Country.}
\Cref{fig:heatmap_person_landmark} (left) illustrates Alignment across Person and Landmark Country. We expected higher performance when the person and landmark originate from the same country, suggesting challenges in cross-cultural representation. However, results vary by country. For instance, the highest alignment occurs when Indian or Vietnamese people visit U.S. landmarks (0.34), comparable to U.S. people at U.S. landmarks (0.33). In contrast, the lowest alignment is observed when Vietnamese people visit Spanish landmarks (0.28). All metrics are detailed in \Cref{sec:intersection}.

\begin{figure}
\centering
\includegraphics[width=0.8\linewidth]{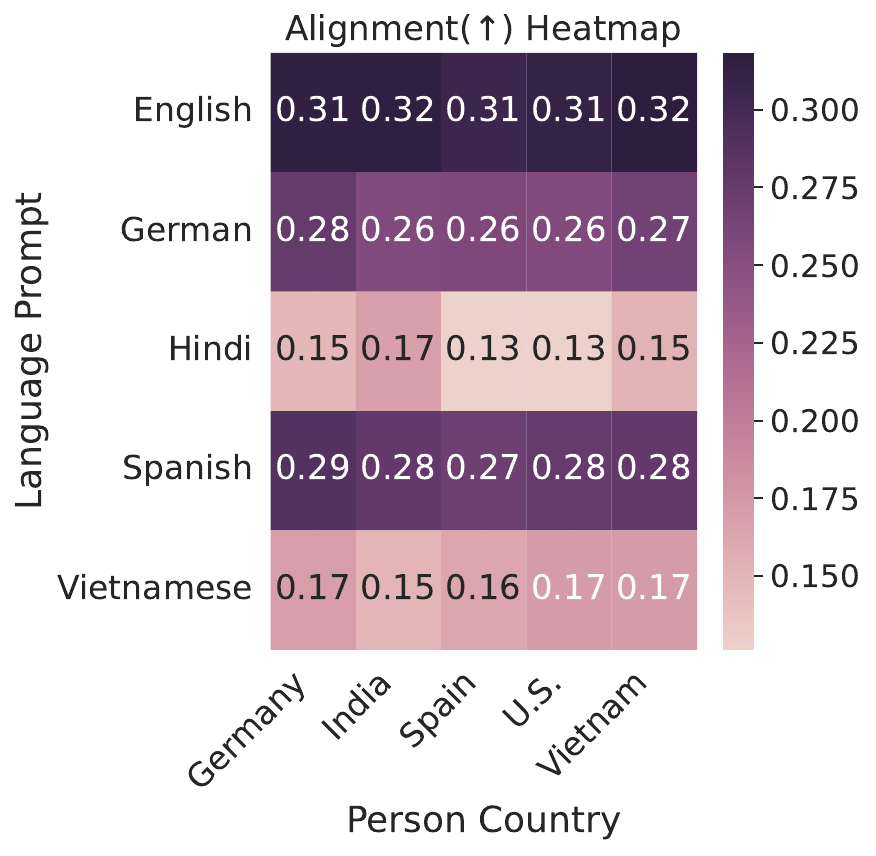}
  \vspace{-0.8em}
\caption{
Alignment with best multilingual model, \texttt{Alt-M}, over image caption language and person country.}
\vspace{-1.5em}
\label{fig:heatmap_person_country_language}
\end{figure}

\noindent\textbf{Language and Country.}
\Cref{fig:heatmap_person_country_language} shows Alignment across Person Country and Caption Language. 
English, Spanish, and Vietnamese captions achieve the highest performance ($\sim$ 0.3) with minimal variation across person countries. However, Hindi captions perform best for Indian people (0.17) and worst for Spanish and U.S. people (0.13). This suggests that, for certain languages, the interaction between caption language and the depicted person’s culture influences Alignment in image generation.

\vspace{-1em}
\subsection{Human Evaluation and Error Analysis}\label{sec:error}
Two annotators evaluate a subset of 300 images, covering all demographics (age, gender, country, landmark) and model settings (Alt-S, Alt-M, Flux-S, Flux-M). They assess the generated images based on three key metrics: Alignment, Quality, and Aesthetics.
Following \citet{lee2024holistic}, Quality is measured in terms of photorealism, while Aesthetics is evaluated based on subject clarity and overall visual appeal.
Annotator agreement is measured using weighted Cohen’s Kappa for ordinal values \cite{Cohen1968WeightedKN}, yielding scores between 0.5 and 0.6 across all three metrics, indicating moderate agreement.
The complete set of human evaluation questions, along with the annotation interface, is detailed in \Cref{sec:human_eval}.

\noindent\textbf{Most Common Errors.}
Across models, errors primarily involve incorrect backgrounds and failures in human rendering. For \texttt{Flux-M}, background inaccuracies are most frequent (38/75), followed by deviations from prompt details and occasional human rendering errors (5/75), such as missing fingers or misplaced cultural markers; landmark errors are comparatively rare (2/75). In contrast, \texttt{Flux-S} exhibits substantially more landmark omissions (15/75) and increased human rendering errors (10/75), particularly for traditional attire and facial features. The \texttt{Alt} models show more severe artifacts overall, with frequent background errors (55/75), pronounced body distortions, and multiplicity errors. While \texttt{Alt-M} reduces culturally related errors (2/75), it still exhibits body distortions (15/75).

\begin{figure*}[htbp]
\centering
  \includegraphics[width=\linewidth]{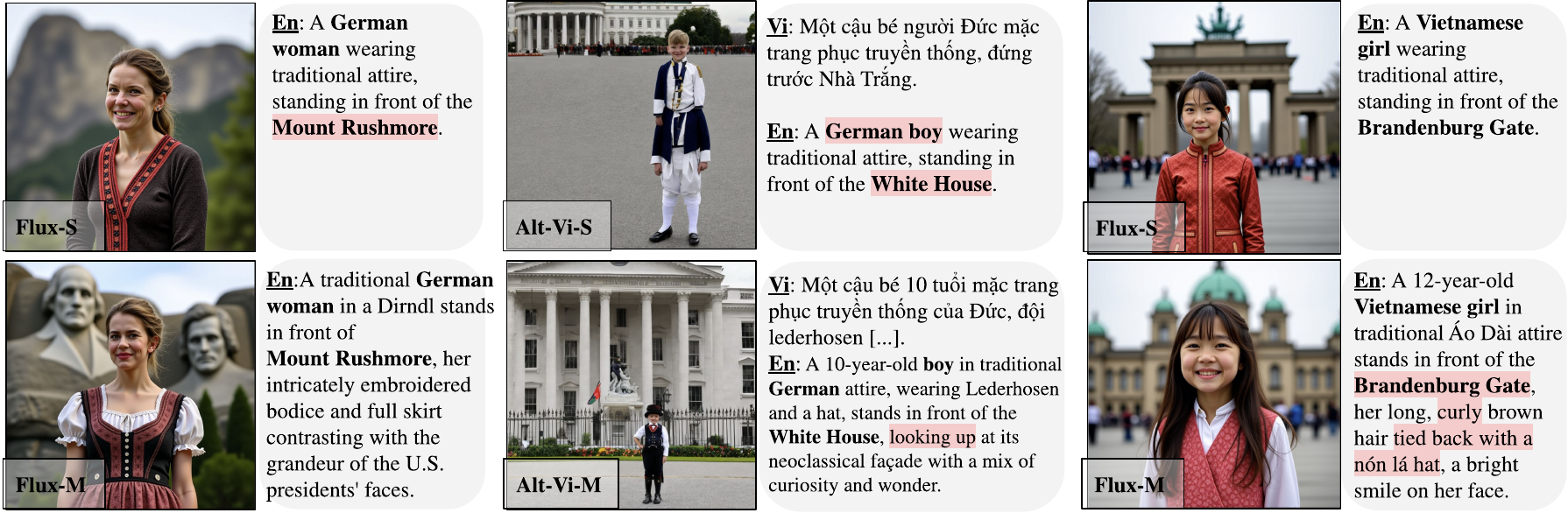}  
  \vspace{-1em}
\caption{ 
Comparison of generated images and captions from multi-agent  (\texttt{Flux-M}, \texttt{Alt-M}) and simple models (\texttt{Flux-S}, \texttt{Alt-S}). The first two columns show where multi-agent models perform better, while the last column shows where simpler models excel. The second column depicts images generated with \textit{Vietnamese} captions using the multilingual model \texttt{Alt} (\texttt{Alt-Vi-S}, \texttt{Alt-Vi-M}). Demographic keywords are \textbf{bolded}, and errors are marked in {\color{red}red}.}
\vspace{-1em}
\label{fig:qualitative.pdf}
\end{figure*}

\subsection{Qualitative Results}\label{sec:quals}

In \Cref{fig:qualitative.pdf}, we compare the images generated by our multi-agent framework (\texttt{Flux-M} and \texttt{Alt-M}) with those from simpler models (\texttt{Flux-S} and \texttt{Alt-S}). The second column presents images generated with Vietnamese captions using the multilingual models (\texttt{Alt-Vi-S}, \texttt{Alt-Vi-M}).
Compared to the simple models, the multi-agent models perform better at generating landmarks and people. However, they still miss important details about people, such as \textit{a person looking up}, \textit{curly hair}, or \textit{hair tied back with a nón lá hat}. Notably, body distortions are more pronounced in the Alt-S model. While the Flux model produces more accurate backgrounds, they tend to be blurrier compared to those in the Alt model.

A manual error analysis of 300 images across all demographics highlights the need for further improvements, particularly in rendering body structures and backgrounds. Models also exhibit systematic tendencies toward stereotypical representations for certain demographics, for instance defaulting to generic or exaggerated traditional attire regardless of age group, or rendering culturally ambiguous poses and accessories. Cross-cultural grounding challenges are further reflected in performance disparities across country-landmark pairs: as shown in Figure 5, Vietnamese people at Spanish landmarks consistently yield the lowest alignment scores (0.28), suggesting that certain cultural combinations remain particularly difficult for current models to represent accurately. Failure modes in multilingual captions are similarly evident in Appendix E.3, where Hindi and Vietnamese captions show increased rendering errors and cultural inaccuracies compared to English. Additional results across demographics are in \Cref{sec:qual}.

\section{Lessons Learned and Actionable Steps}

Our study surfaces insights into the challenges of \emph{multicultural text-to-image generation} and highlights several directions for improving cultural grounding, demographic coverage, and evaluation practices in future models.

\noindent\textbf{Richer Cultural and Demographic Captions.}
Multicultural generation benefits from richer prompts that clearly articulate cultural context, demographic attributes, and landmark-specific details.
By integrating diverse perspectives through collaboration, multi-agent models enhance alignment, aesthetics, quality, and knowledge (\Cref{sec:multiagent}). Future research should focus on refining multi-agent frameworks to further enhance alignment and representational diversity. 
Our work can also be extend to evaluate a broader range of cultural interactions—such as social activities, rituals, and everyday practices—to better assess reasoning and action-based image generation.

\noindent\textbf{Multilingual Support Remains a Bottleneck.}
We observe systematic performance gaps between English and non-English prompts, with English-language generations consistently achieving higher alignment scores (\Cref{fig:ablations}e). These disparities point to limitations in current multilingual training and evaluation practices. Improving multilingual coverage—both in training data and model architectures—is essential for achieving equitable performance across languages and cultures.

\noindent\textbf{Develop Better Evaluation Metrics.}
Automated metrics do not always align with qualitative judgments in multicultural settings, particularly when visually plausible context inflates scores despite incorrect culturally salient elements (e.g., accurate surroundings but an incorrect landmark; \Cref{sec:error}). More reliable evaluation requires metrics that place greater emphasis on landmark identity, demographic attributes, and compositional correctness. Targeted automated measures, complemented by human evaluation, remain essential for accurately assessing multicultural image generation.

\noindent\textbf{Explore Stronger and More Diverse Baselines.}
MosAIG establishes multi-agent collaboration as one principled strategy for multicultural prompt composition. Future work should compare against complementary approaches such as single-agent chain-of-thought (CoT) prompt expansion and template-based demographic augmentation, which would help disentangle the specific contribution of multi-agent decomposition from the general benefit of richer prompt specification.

\noindent\textbf{Finer-Grained Compositional Evaluation.}
While CLIPScore provides a scalable alignment proxy, future work should incorporate more direct evaluation approaches such as detection-based attribute binding checks, explicit landmark classification over generated images, and LLM-as-judge VQA methods. Such measures would provide finer-grained diagnostics of cultural and demographic grounding, particularly for compositional correctness across person-landmark pairs.

\section{Conclusion}
In this paper, we introduce \emph{multicultural text-to-image generation} as a new task for evaluating how models depict people and landmarks from different cultural backgrounds within a single image. We release \textsc{MosAIG} the first benchmark for this setting, comprising 9,000 images across five countries, three age groups, two genders, 25 historical landmarks, and five languages.
Our automated and human evaluations reveal substantial variation across demographics, languages, and cultural configurations, highlighting persistent gaps in multilingual and cross-cultural generation. Overall, our findings emphasize the importance of explicit cultural and demographic grounding for improving image quality, factual correctness, and representational consistency.
We release our dataset and evaluation framework to support multicultural text-to-image generation:
\url{https://github.com/AIM-SCU/MosAIG}.

\section*{Limitations and Ethical Considerations}

\paragraph{Scope of Demographic Coverage.}
Our study considers a limited set of demographic attributes, focusing on binary gender categories, three coarse age groups (child, adult, elder), and five countries and languages. These choices necessarily simplify the rich diversity of gender identities, life stages, and cultural experiences, and limit our ability to assess performance for underrepresented communities. We view this design as a proof of concept for studying multicultural image generation in a controlled setting. Importantly, our dataset and pipeline are fully open-source and designed to be easily extended to additional countries, languages, age groups, and gender identities in future work.

\paragraph{Challenges in Modeling Cultural Identity.}
Our approach relies on structured prompts to approximate cultural and demographic context, but identity is inherently complex and cannot be fully captured through high-level attributes such as nationality, language, age, or gender alone~\cite{saha-etal-2025-meta}. Defining culture primarily through national affiliation risks overlooking substantial intra-cultural variation and lived experience. Future work should incorporate richer contextual dimensions—such as historical background, social practices, and personal narratives—to enable more nuanced and authentic representations.

\paragraph{Limitations of Automated Alignment Metrics.}
We rely on CLIPScore as a scalable, reference-free measure of text--image alignment, but this metric has several limitations. Its coarse-grained contrastive training makes it insensitive to fine-grained compositional errors, such as incorrect spatial relationships or misattributed attributes, and it may assign high scores to images that contain the correct objects in incorrect configurations~\cite{hessel2021clipscore}. CLIPScore is also largely insensitive to word order and linguistic phenomena such as negation, and exhibits biases toward salient or centrally positioned objects, reducing its reliability for complex, multi-object prompts~\cite{Castro2023ScalablePA, abbasi2025}. We therefore encourage future work to complement CLIPScore with image-based classifiers and targeted evaluation methods that more directly assess visual grounding and demographic fidelity~\cite{hu2023tifa}.

\paragraph{Limitations of Human Evaluation Coverage.}
Our human evaluation is conducted on a carefully stratified sample of 300 images covering all demographic categories, model settings, and languages, ensuring representative coverage across the experimental matrix. Expanding human evaluation to a larger sample would provide additional validation of demographic specific findings, particularly for fine-grained combinations of age, gender, country, and landmark. This is a non-trivial undertaking, as it requires annotators proficient in all five languages covered in our benchmark, and we highlight this as a valuable direction for future work to build upon.


\bibliography{custom}

\appendix

\section{Appendix}
\label{sec:appendix}

\section{Data}\label{sec:model}

\begin{table*}[h]
    \centering
    \begin{tabular}{|c|c|c|l|}
        \hline
        \textbf{Age} & \textbf{Gender} & \textbf{Country} & \textbf{Landmark} \\ 
        \hline
        \multirow{25}{*}{Child/ Adult/ Elder} & \multirow{25}{*}{Female/Male} & \multirow{5}{*}{Germany} & Cologne Cathedral \\ 
        & & & Reichstag Building \\ 
        & & & Neuschwanstein Castle \\ 
        & & & Brandenburg Gate \\ 
        & & & Holocaust Memorial \\ 
        \cline{3-4}
        & & \multirow{5}{*}{India} & Taj Mahal \\ 
        & & & Lotus Temple \\ 
        & & & Gateway of India \\ 
        & & & India Gate \\ 
        & & & Charminar \\ 
        \cline{3-4}
         &  & \multirow{5}{*}{Spain} & Sagrada Familia \\ 
        & & & Alhambra \\ 
        & & & Guggenheim Museum \\ 
        & & & Roman Theater of Cartagena \\ 
        & & & Royal Palace of Madrid \\ 
        \cline{3-4}
        & & \multirow{5}{*}{U.S.} & White House \\ 
        & & & Statue of Liberty \\ 
        & & & Mount Rushmore \\ 
        & & & Golden Gate Bridge \\ 
        & & & Lincoln Memorial \\ 
        \cline{3-4}
         &  & \multirow{5}{*}{Vietnam} & Meridian Gate of Huế \\ 
        & & & Independence Palace \\ 
        & & & One Pillar Pagoda \\ 
        & & & Ho Chi Minh Mausoleum \\ 
        & & & Thien Mu Pagoda \\ 
        \hline
    \end{tabular}
    \caption{Demographics Overview: 3 Age groups, 2 Genders, 5 Countries, and 25 Landmarks}
    \label{tab:age_gender_country_landmark}
\end{table*}

\section{Multicultural Image Generation}\label{sec:model}

\subsection{Implementation Details}

The Summarizer Agent and each Social Agent are initialized as different instances of a LLaMA model\footnote{\url{https://huggingface.co/meta-llama/Llama-3.1-8B}}~\cite{touvron2023llama}.
The Moderator Agent is a predefined function call. The agent conversation uses the CrewAI framework to establish an iterative feedback loop\footnote{https://www.crewai.com/open-source}.
The implementation was carried out using an NVIDIA V100 GPU (32GB). More details can be found in \Cref{sec:model}.

The multi-agent configuration processed 750 base prompts in approximately 45 minutes, while additional language variants (3,750 prompts in total) required 75 minutes using the Google Translation API.
Two models—Flux and Alt-Diffusion—were used for image generation:
Flux produced 750 images (768×768 pixels) in 2.5 hours with the settings: guidance scale: 4, inference steps: 30, seed: 11, averaging roughly 12 seconds per image.
Alt-Diffusion was configured with the settings: guidance scale: 11, inference steps: 110, seed: 11000, and processed 3,750 images of the same resolution in 16 hours, averaging about 15 seconds per image.
All processing times accounted for overhead related to model loading and image saving, ensuring consistency in image resolution (768×768 pixels) across both models.

\begin{figure*}
\centering
\includegraphics[width=\linewidth]{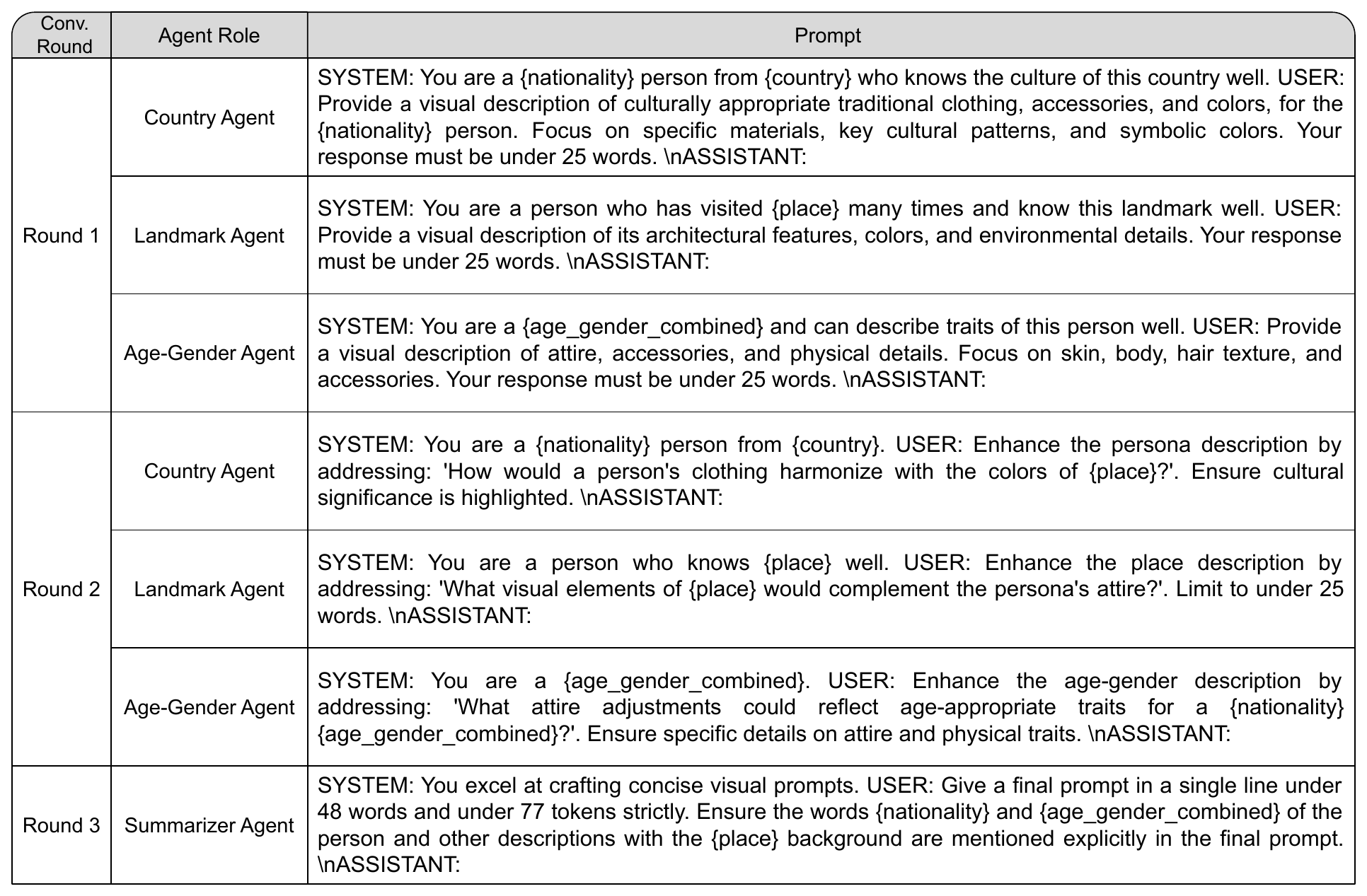}
\caption{Our Multi-agent Framework Prompts}

\label{fig:prompts}
\end{figure*}

\section{Human Evaluation and Error Analysis}\label{sec:human_eval}

We rely on human annotators to assess a sample of the generated images based on three key metrics: Alignment, Quality, and Aesthetics.
Following ~\citet{lee2024holistic}, Quality is evaluated in terms of photorealism, while Aesthetics is assessed based on subject clarity and overall visual appeal. The complete set of human evaluation questions is outlined below.
Annotators are provided with definitions (\Cref{tab:guidelines_human_eval}) and corresponding questions to guide their assessments. To determine whether the generated images meet their expectations, we ask annotators to rate them using a 5-point Likert scale.

\begin{table*}[h]
\centering
\begin{tabular}{c|p{12cm}}
\toprule
Aspect & Definition \\
\midrule
Alignment    &   Is the image semantically correct given the text (text-image alignment)?        \\
Quality    &    Do the generated images look like real photographs?       \\
Aesthetic    &    Is the image aesthetically pleasing?       \\
Fairness    &      Does the model exhibit performance disparities across social groups (e.g., gender, dialect)     \\
Knowledge    &   Does the model have knowledge about the world or domains?        \\
\bottomrule
\end{tabular}%
\caption{Evaluation Aspects of Text-to-Image Models}
\label{tab:guidelines_human_eval}
\end{table*}

\paragraph{Alignment.}
We ask the annotators to rate how well the image matches the description.

\textbf{How well does the image match the description?}
\begin{enumerate}
    \item  Does not match at all
    \item  Has significant discrepancies
    \item  Has several minor discrepancies
    \item  Has a few minor discrepancies
    \item  Matches exactly
\end{enumerate}

\paragraph{Quality.}
We ask the annotators to rate how photorealistic the generated images are.

\textbf{Determine if the following image is AI-generated or real.}
\begin{enumerate}
    \item  AI-generated photo.
    \item  Probably an AI-generated photo, but photorealistic.
    \item  Neutral.
    \item  Probably a real photo, but with irregular textures and shapes.
    \item  Real photo.
\end{enumerate}

\paragraph{Aesthetics.}
To evaluate the overall aesthetics, we ask annotators to provide a holistic assessment of the image's visual appeal by rating its aesthetic quality.

\textbf{How aesthetically pleasing is the image?}
\begin{enumerate}
    \item I find the image ugly.
    \item The image has a lot of flaws, but it's not completely unappealing.
    \item I find the image neither ugly nor aesthetically pleasing.
    \item  The image is aesthetically pleasing and is nice to look at.
    \item The image is aesthetically stunning. I can look at it all day.
\end{enumerate}

\begin{figure*}
\centering
\includegraphics[width=0.9\linewidth]{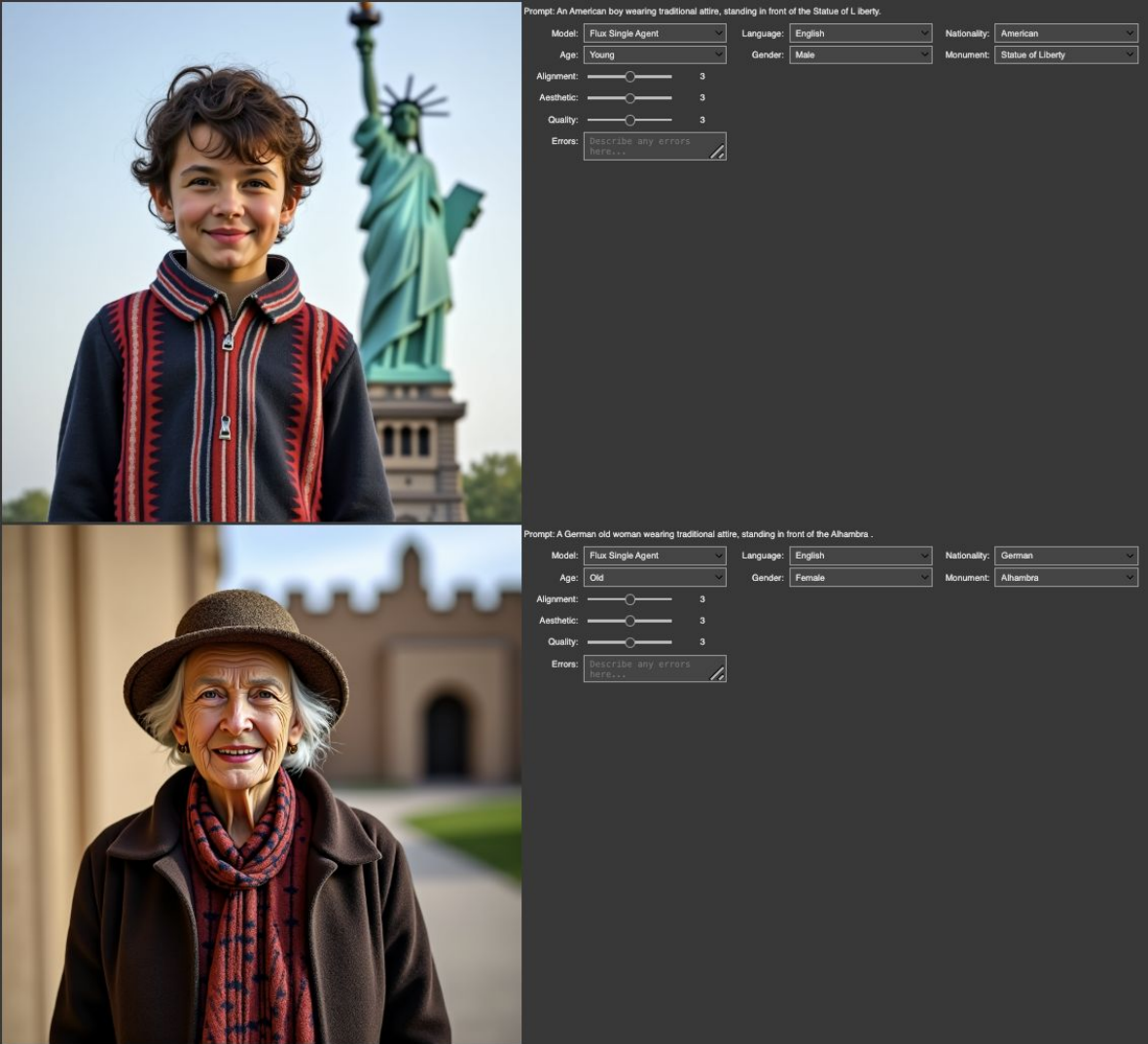}
\caption{Human Annotation Interface for manually evaluating the images across all models.}
\label{fig:annotation_pipeline}
\end{figure*}

\newpage
\section{Results}\label{sec:results}

\newpage


\begin{figure}
\centering
\includegraphics[width=\linewidth]{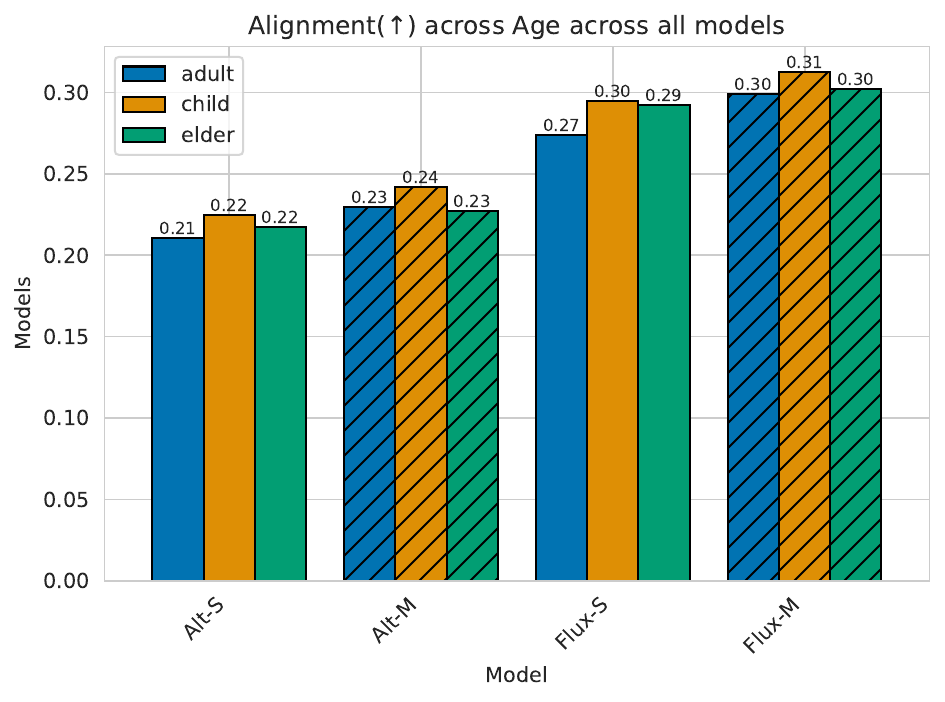}
\vspace{-1em}
\label{fig:all_Age_Alignment}
\end{figure}

\begin{figure}
\centering
\includegraphics[width=\linewidth]{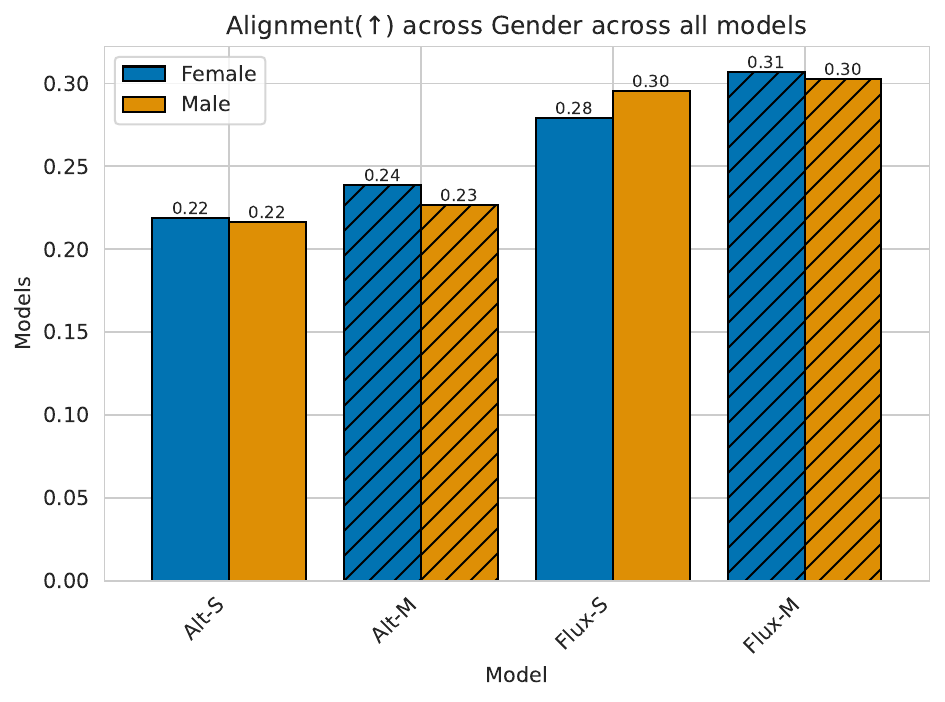}
\vspace{-1em}
\label{fig:all_Gender_Alignment}
\end{figure}

\begin{figure}
\centering
\includegraphics[width=\linewidth]
{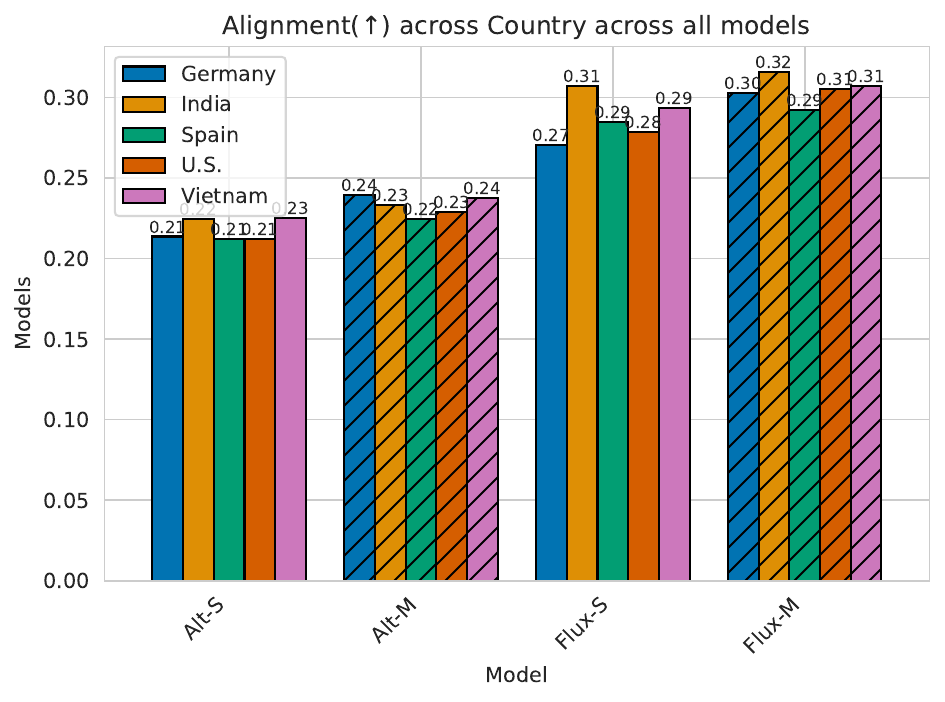}
\vspace{-1em}
\label{fig:all_Country_Alignment}
\end{figure}

\begin{figure}
\centering
\includegraphics[width=\linewidth]{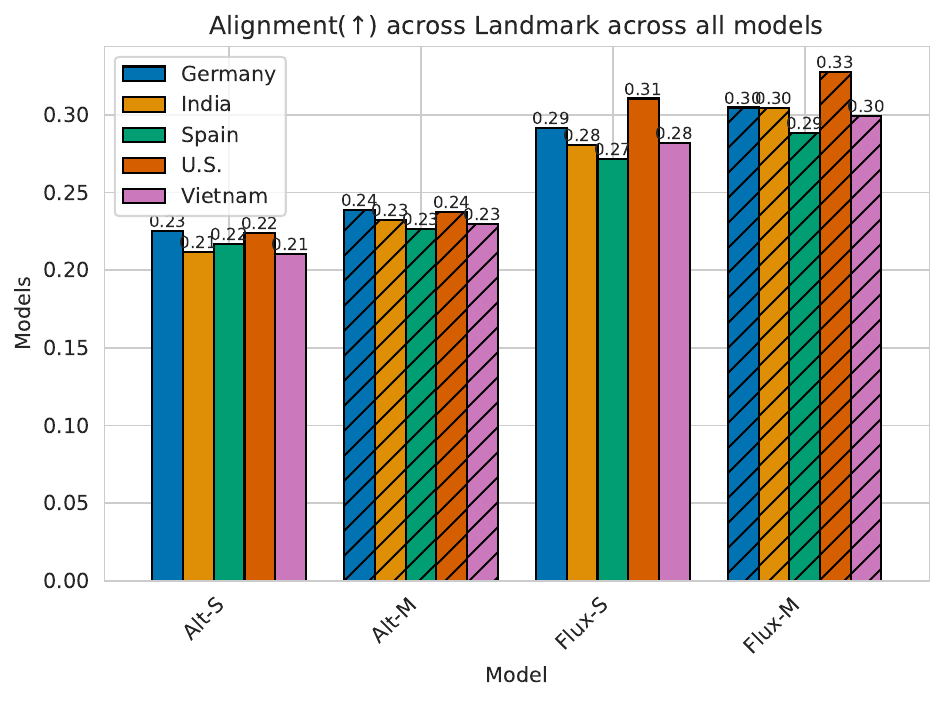}
\vspace{-1em}
\label{fig:all_Landmark_Alignment}
\end{figure}


\begin{figure}
\centering
\includegraphics[width=\linewidth]{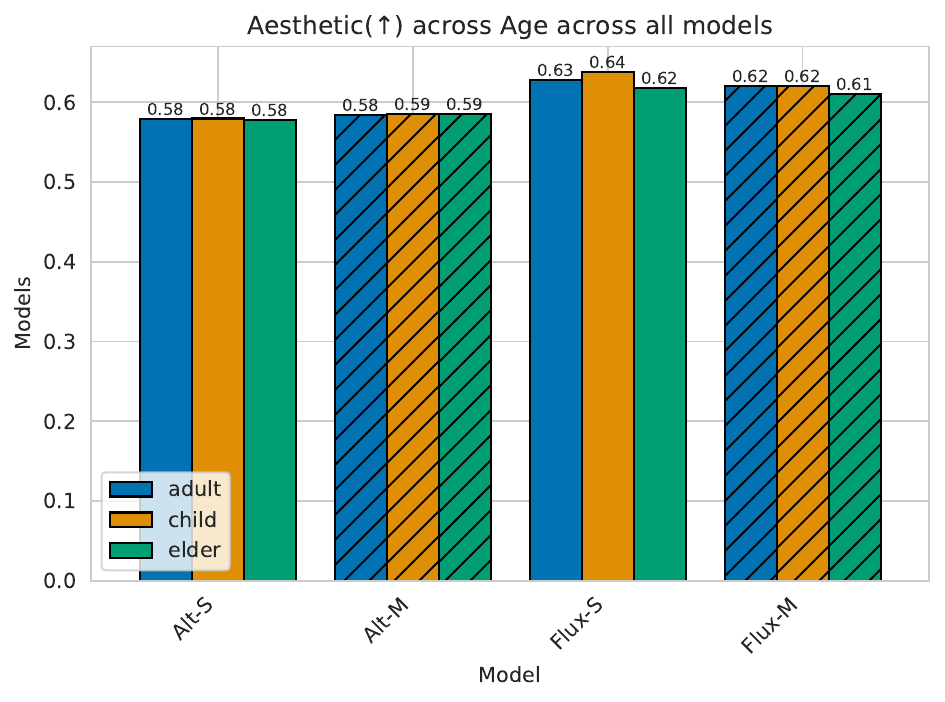}
\vspace{-1em}

\label{fig:all_Age_Aesthetic}
\end{figure}

\begin{figure}
\centering
\includegraphics[width=\linewidth]{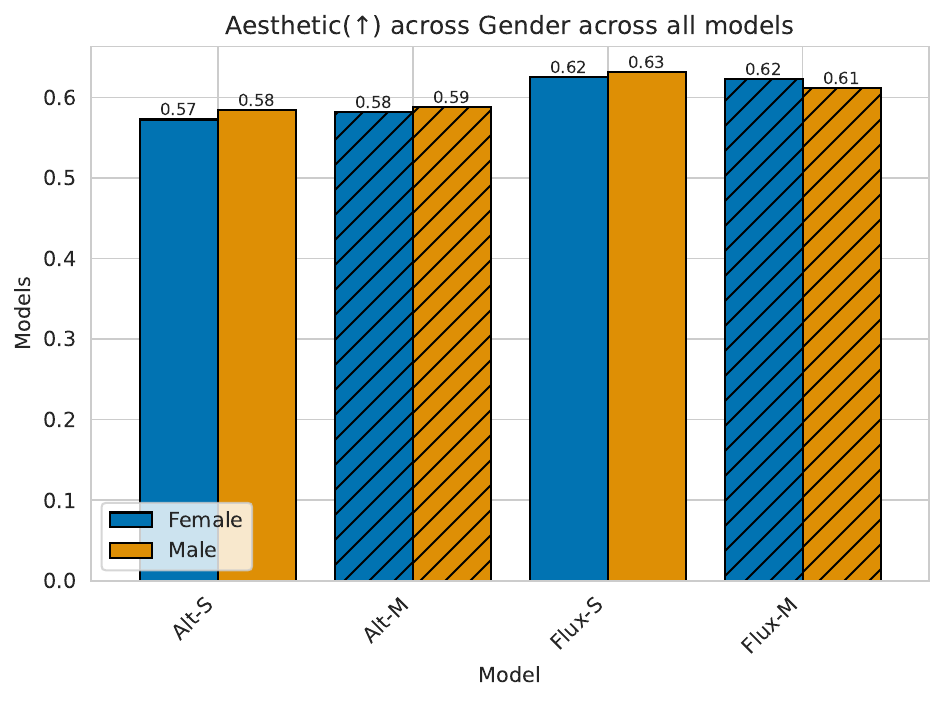}
\vspace{-1em}
\label{fig:all_Gender_Aesthetic}
\end{figure}

\begin{figure}
\centering
\includegraphics[width=\linewidth]{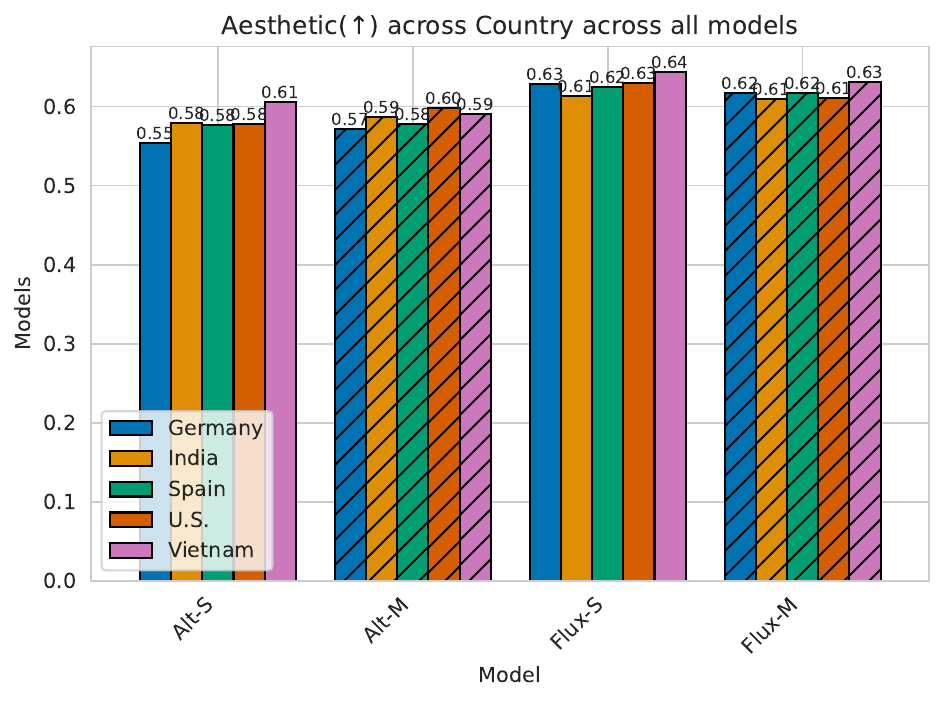}
\vspace{-1em}
\label{fig:all_Country_Aesthetic}
\end{figure}

\begin{figure}
\centering
\includegraphics[width=0.9\linewidth]{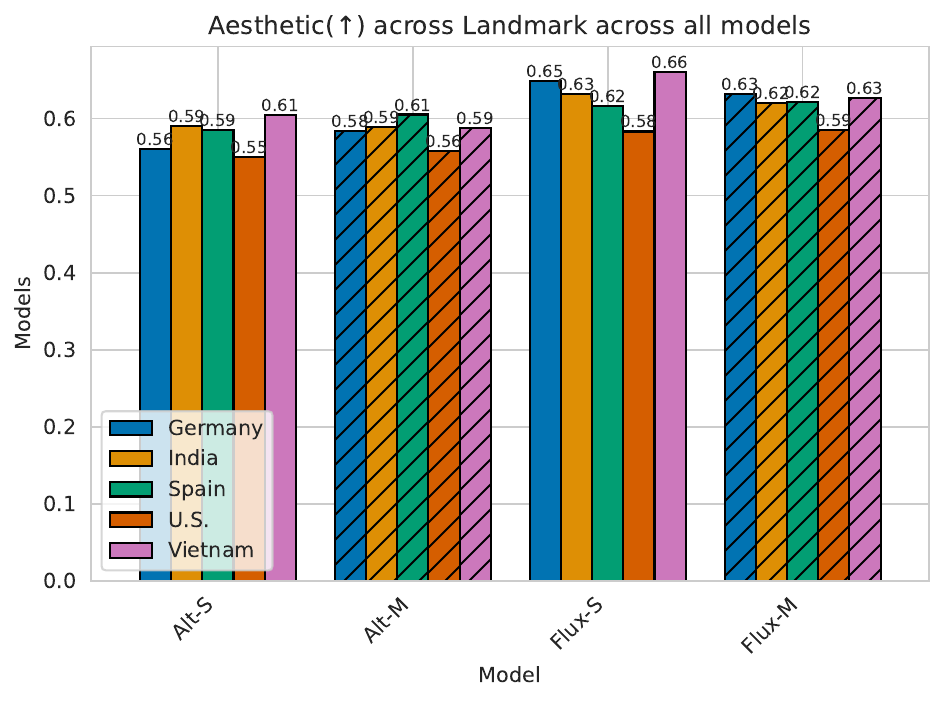}
\vspace{-1em}
\label{fig:all_Landmark_Aesthetic}
\end{figure}

\begin{figure}
\centering
\includegraphics[width=\linewidth]{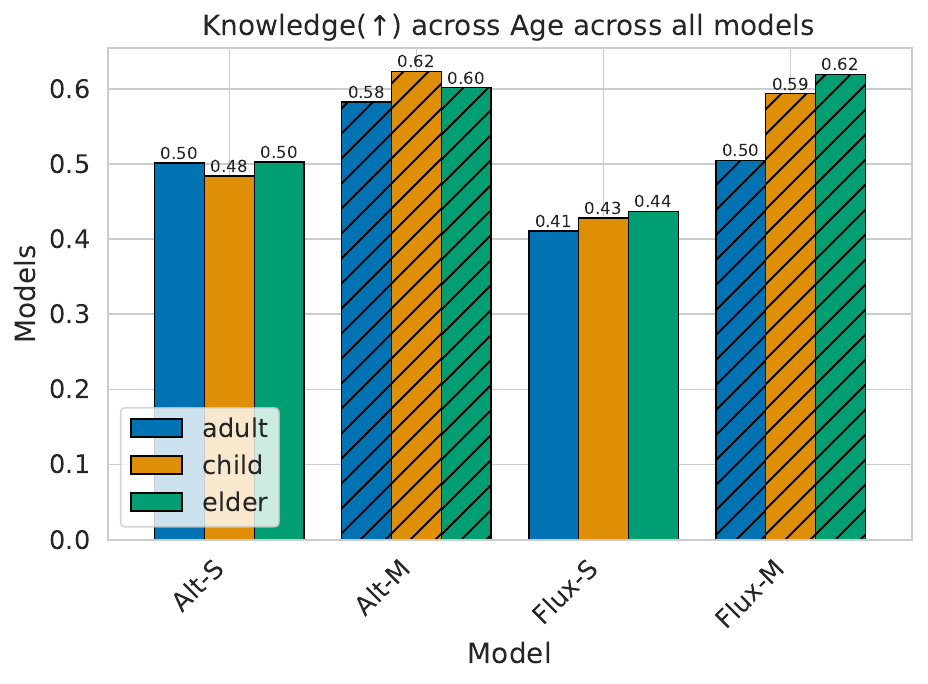}
\vspace{-1em}
\label{fig:all_Age_Knowledge}
\end{figure}

\begin{figure}
\centering
\includegraphics[width=\linewidth]{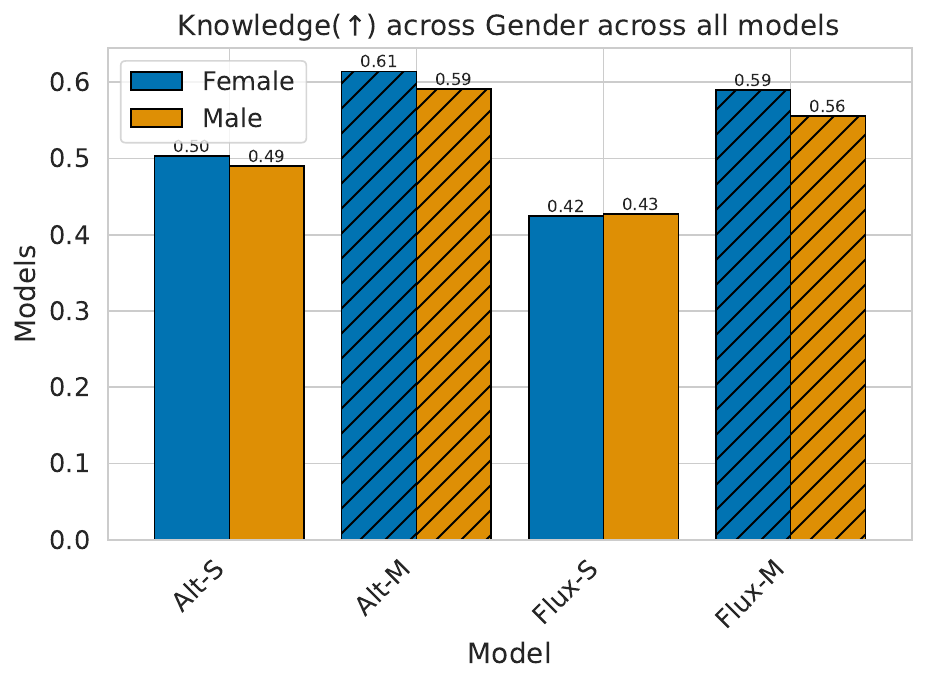}
\vspace{-1em}
\label{fig:all_Gender_Knowledge}
\end{figure}

\begin{figure}
\centering
\includegraphics[width=\linewidth]{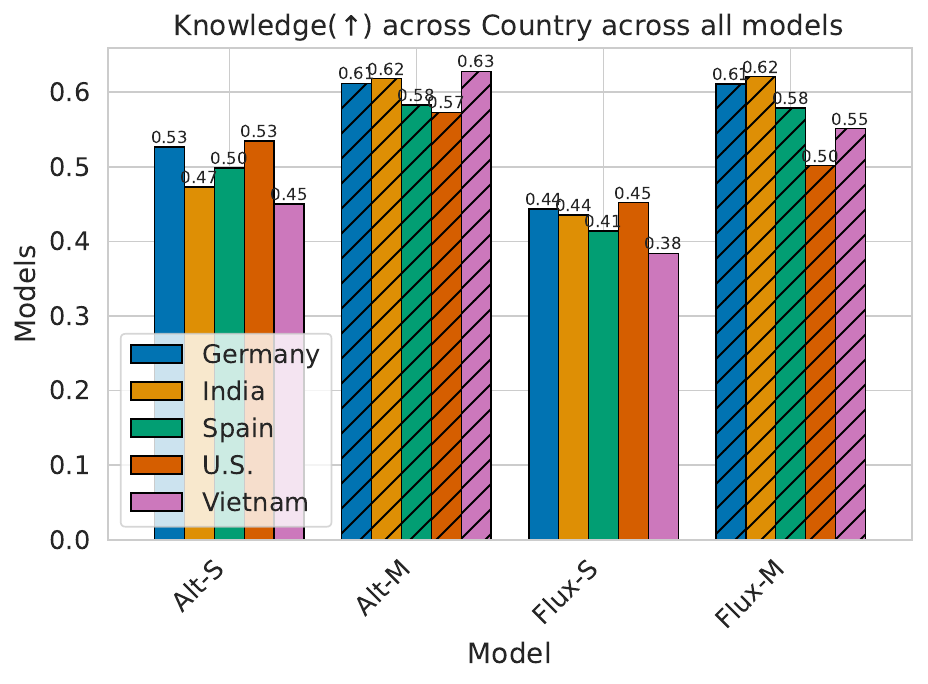}
\vspace{-1em}
\label{fig:all_Country_Knowledge}
\end{figure}

\begin{figure}
\centering
\includegraphics[width=\linewidth]
{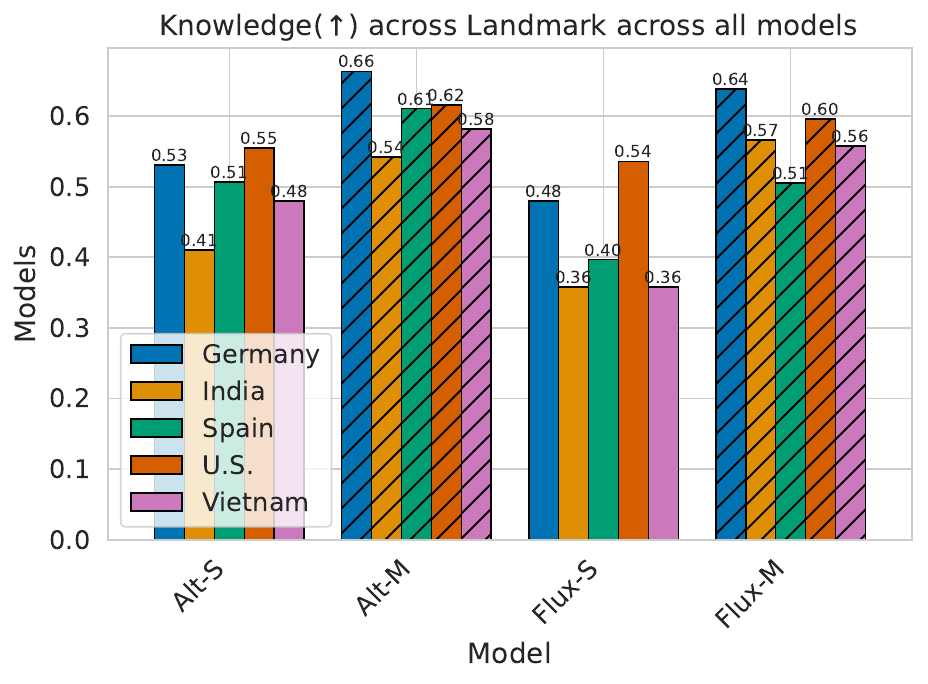}
\vspace{-1em}
\label{fig:all_Landmark_Knowledge}
\end{figure}

\begin{figure}
\centering
\includegraphics[width=\linewidth]{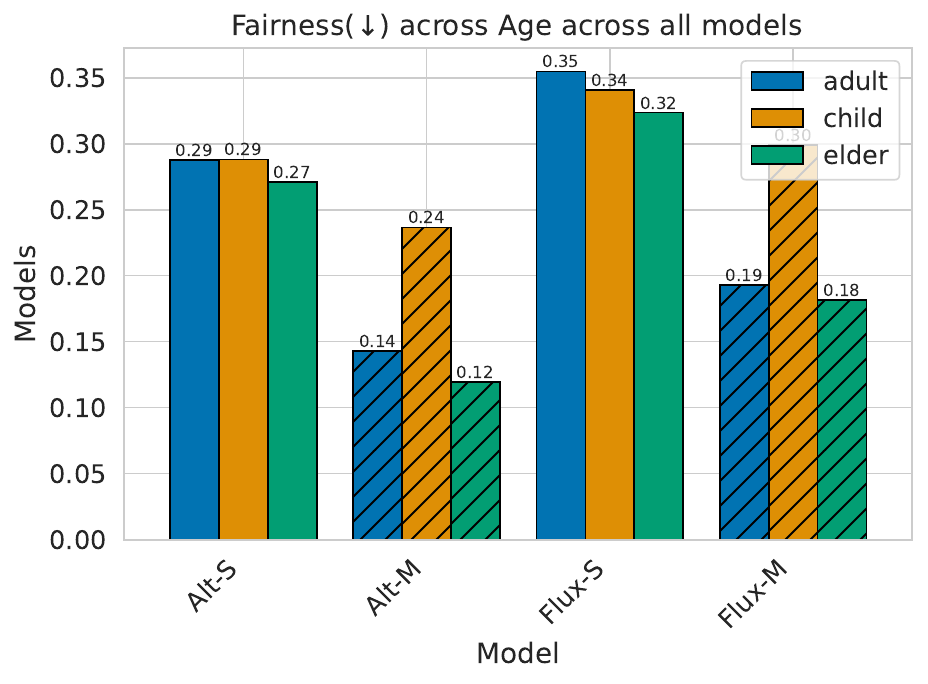}
\label{fig:all_Age_Fairness}
\end{figure}

\begin{figure}
\centering
\includegraphics[width=\linewidth]{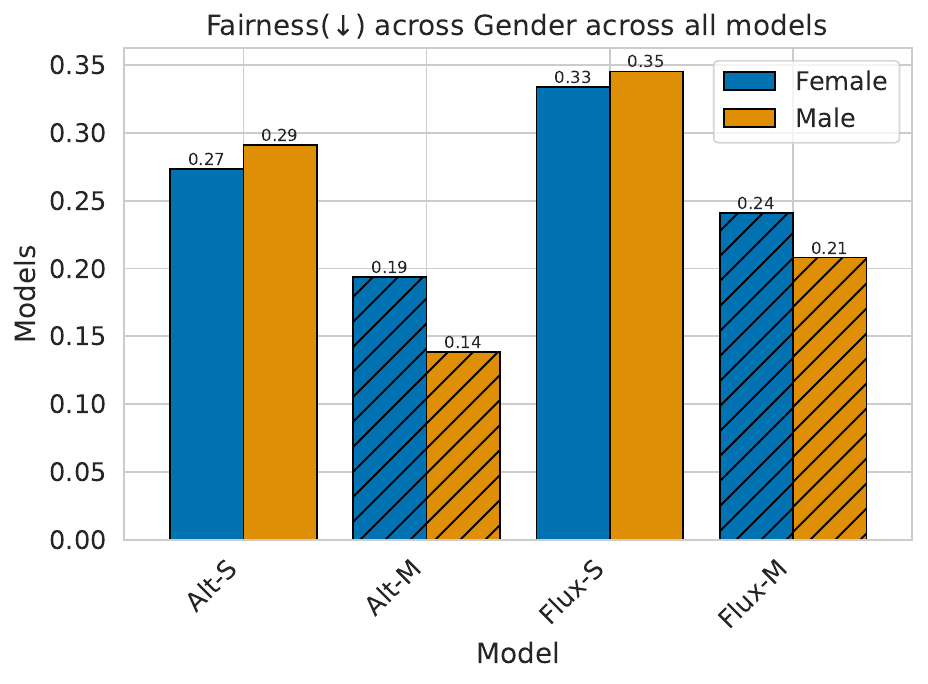}
\vspace{-1em}
\label{fig:all_Gender_Fairness}
\end{figure}

\begin{figure}
\centering
\includegraphics[width=\linewidth]{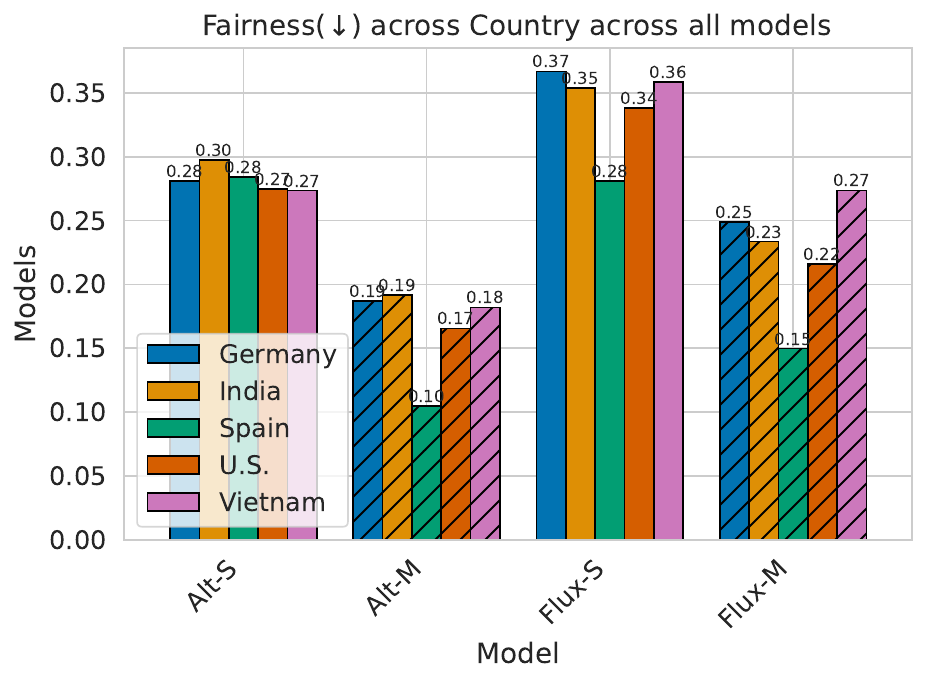}
\vspace{-1em}
\label{fig:all_Country_Fairness}
\end{figure}

\begin{figure}
\centering
\includegraphics[width=\linewidth]{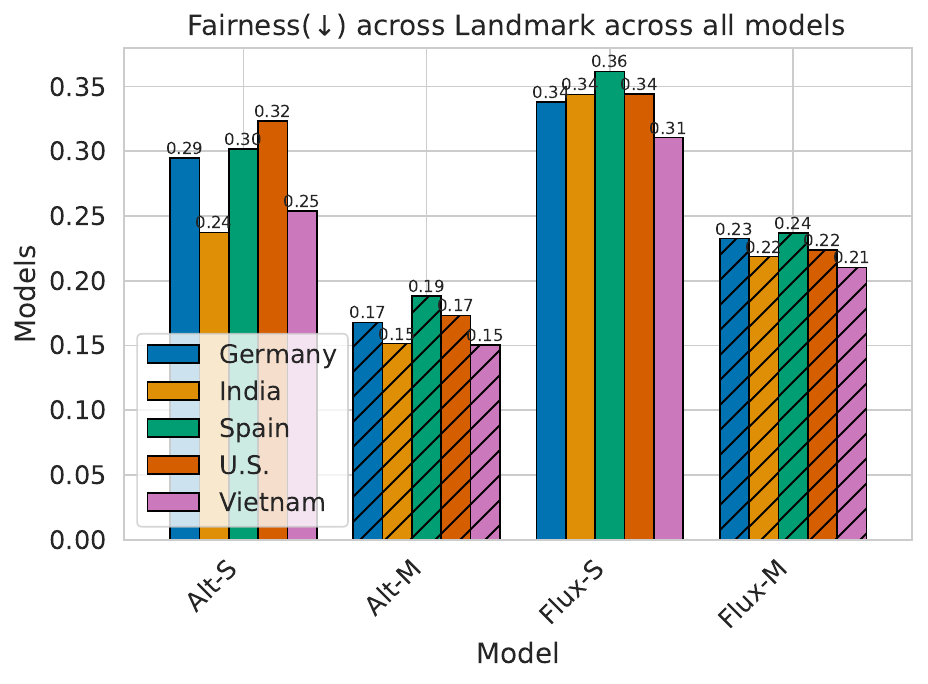}
\vspace{-1em}
\label{fig:all_Landmark_Fairness}
\end{figure}

\newpage
\subsection{Intersectionality}\label{sec:intersection}


\begin{figure}[htbp]
\centering
\includegraphics[width=0.8\linewidth]{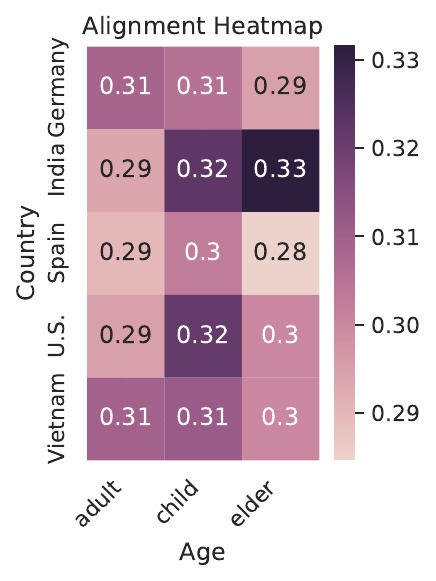}
\label{fig:heatmap_country_age}
\end{figure}

\begin{figure}[htbp]
\centering
\includegraphics[width=0.8\linewidth]{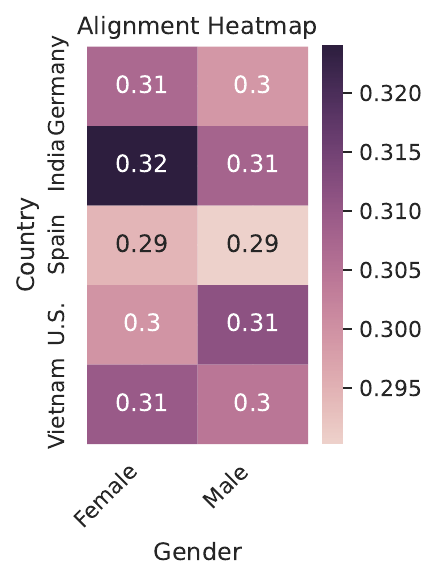}
\label{fig:heatmap_gender_country}
\end{figure}

\begin{figure}[htbp]
\centering
\includegraphics[width=0.8\linewidth]{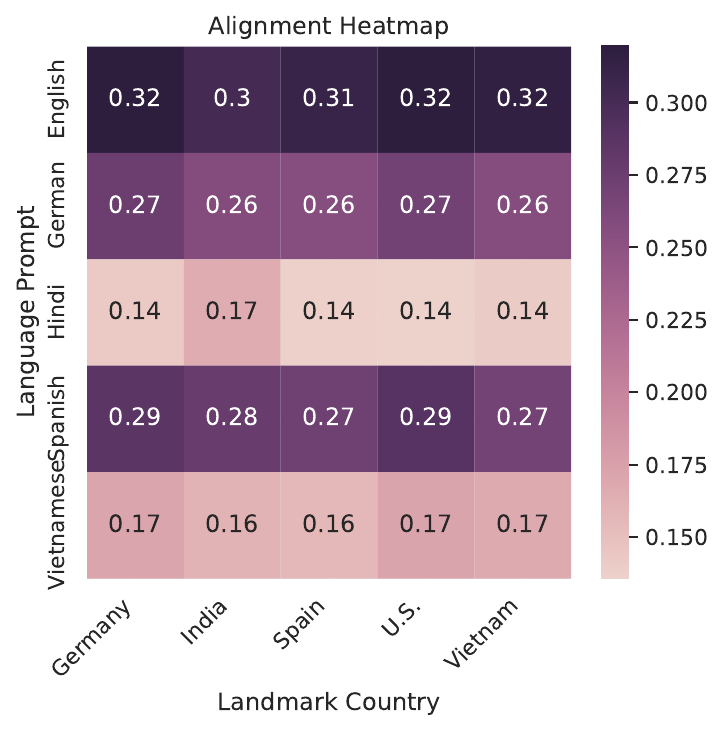}
\label{fig:heatmap_landmark_country_language}
\end{figure}


\begin{figure}[htbp]
\centering
\includegraphics[width=0.8\linewidth]{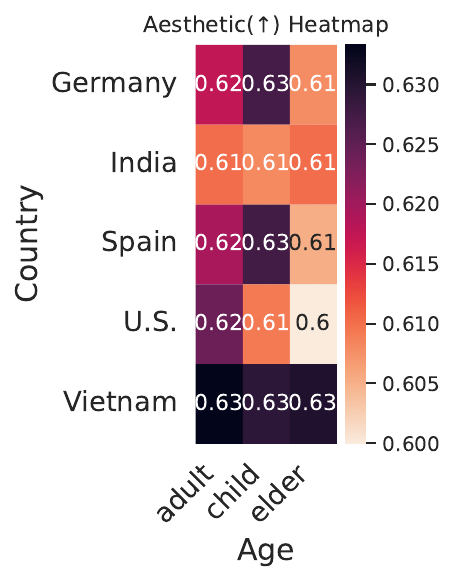}
\label{fig:heatmap_country_age}
\end{figure}

\begin{figure}[htbp]
\centering
\includegraphics[width=0.8\linewidth]{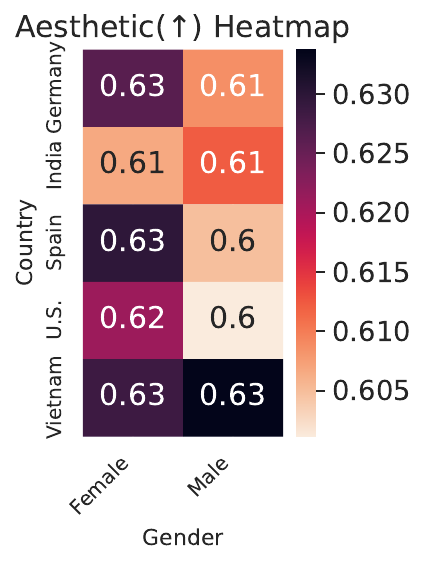}
\label{fig:heatmap_gender_country}
\end{figure}

\begin{figure}[htbp]
\centering
\includegraphics[width=0.8\linewidth]{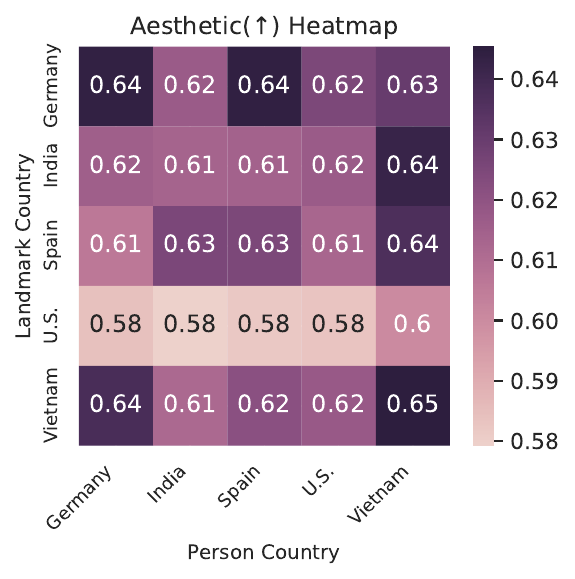}
\label{fig:heatmap_landmark_country_language}
\end{figure}

\begin{figure}[htbp]
\centering
\includegraphics[width=0.8\linewidth]{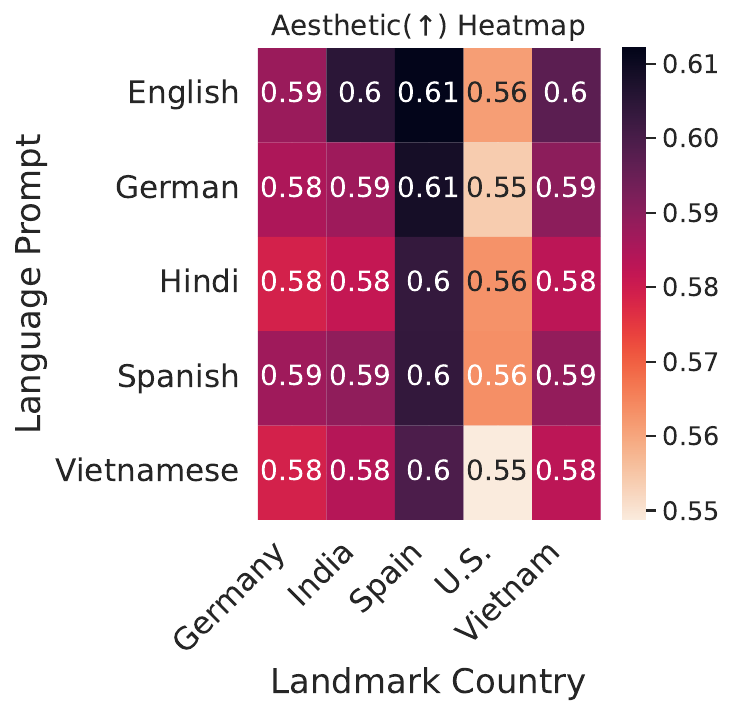}
\label{fig:heatmap_landmark_country_language}
\end{figure}

\begin{figure}[htbp]
\centering
\includegraphics[width=0.8\linewidth]{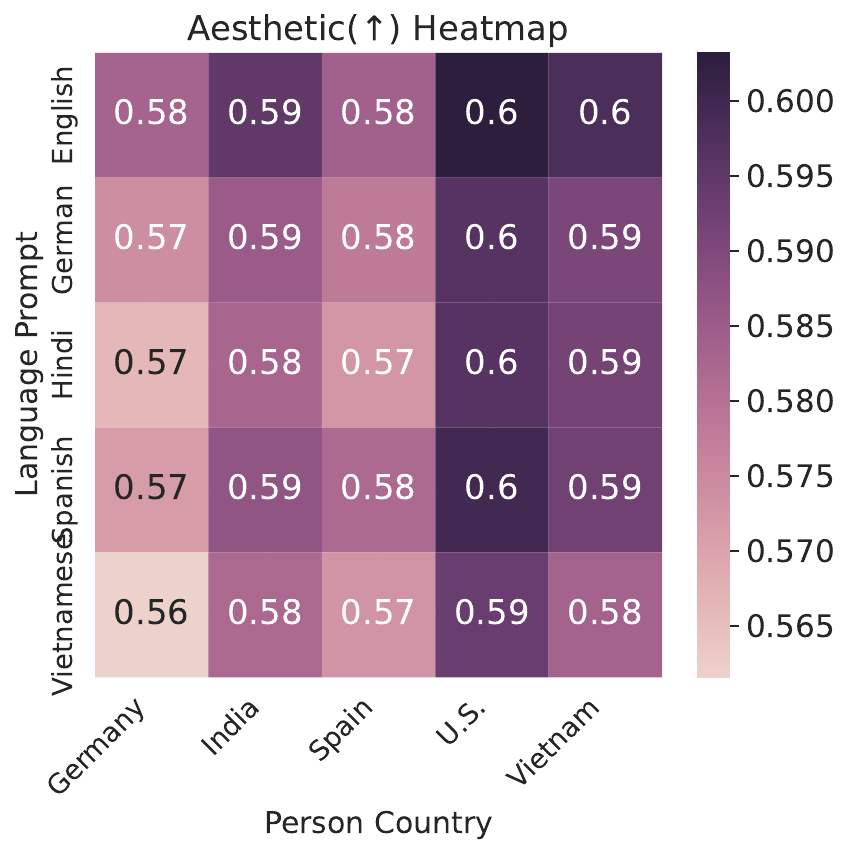}
\label{fig:heatmap_landmark_country_language}
\end{figure}


\begin{figure}[htbp]
\centering
\includegraphics[width=0.8\linewidth]{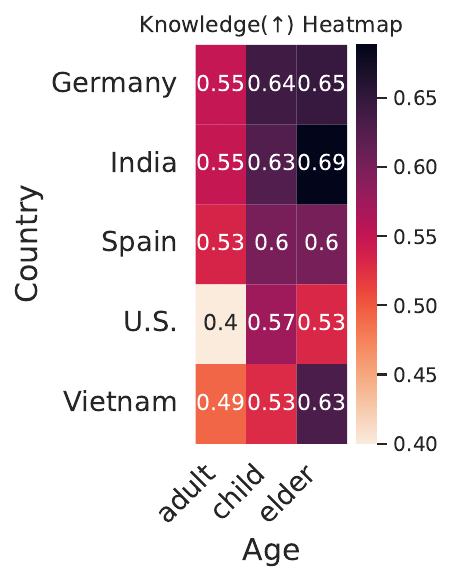}
\label{fig:heatmap_country_age}
\end{figure}

\begin{figure}[htbp]
\centering
\includegraphics[width=0.8\linewidth]{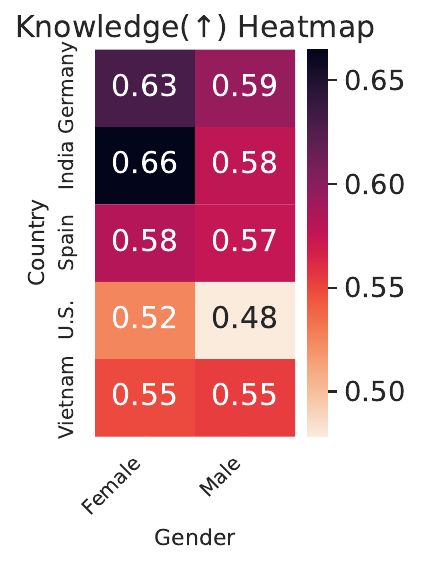}
\label{fig:heatmap_gender_country}
\end{figure}

\begin{figure}[htbp]
\centering
\includegraphics[width=0.8\linewidth]{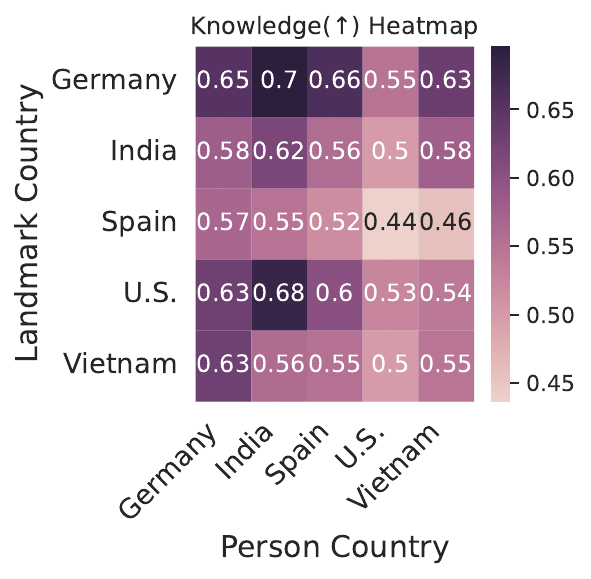}
\label{fig:heatmap_landmark_country_language}
\end{figure}

\begin{figure}[htbp]
\centering
\includegraphics[width=0.8\linewidth]{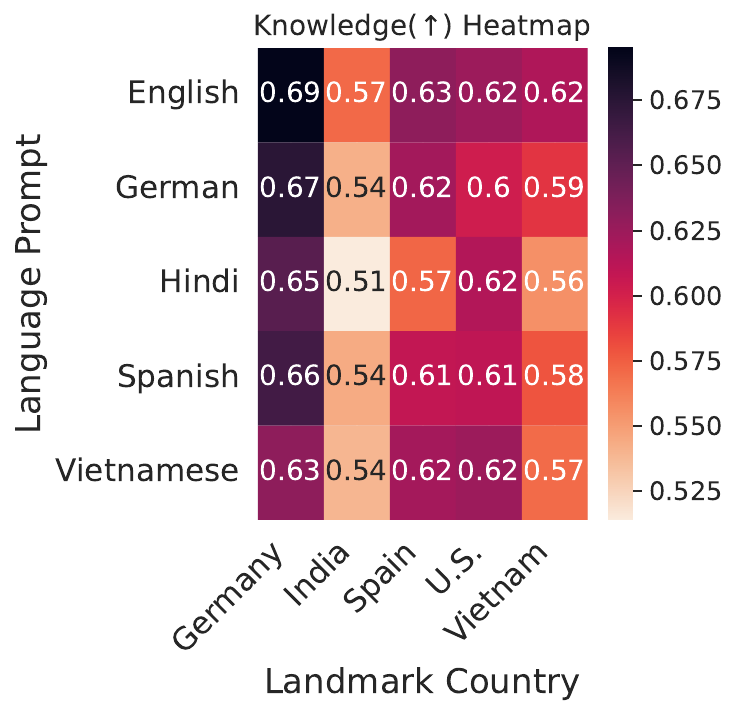}
\label{fig:heatmap_landmark_country_language}
\end{figure}

\begin{figure}[htbp]
\centering
\includegraphics[width=0.8\linewidth]{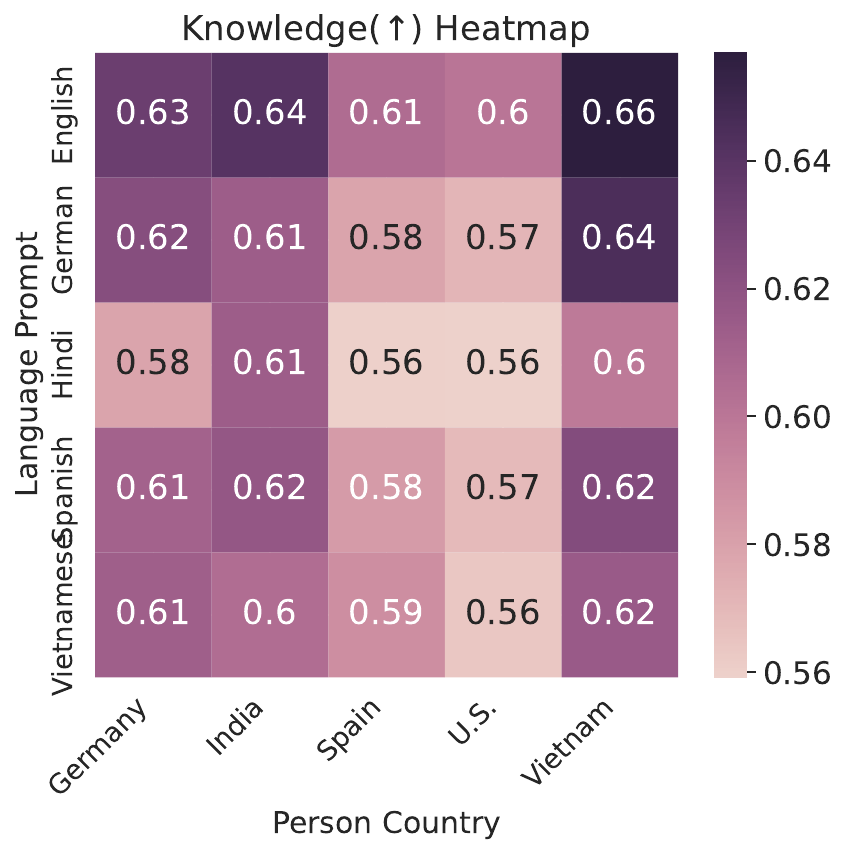}
\label{fig:heatmap_landmark_country_language}
\end{figure}


\begin{figure}[htbp]
\centering
\includegraphics[width=0.8\linewidth]{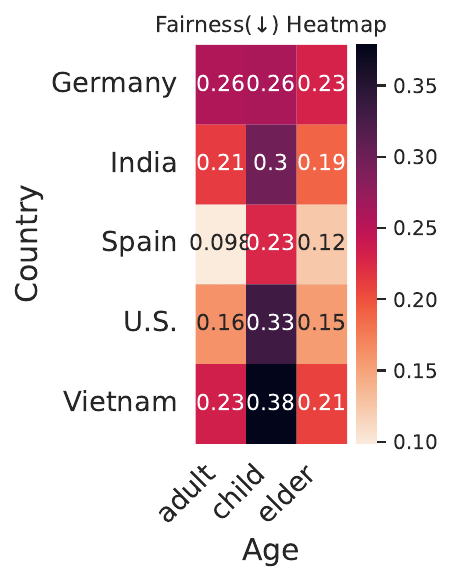}
\label{fig:heatmap_country_age}
\end{figure}

\begin{figure}[htbp]
\centering
\includegraphics[width=0.8\linewidth]{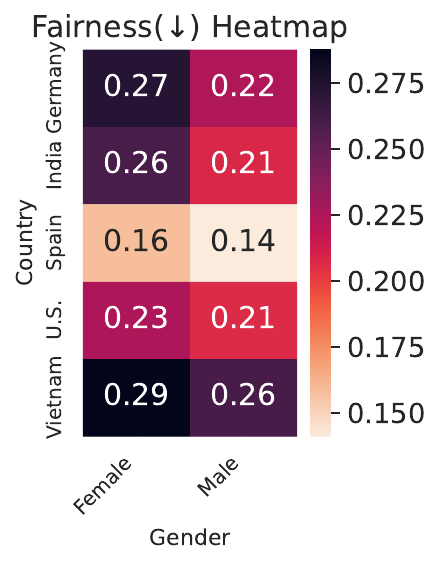}
\label{fig:heatmap_gender_country}
\end{figure}

\begin{figure}[htbp]
\centering
\includegraphics[width=0.8\linewidth]{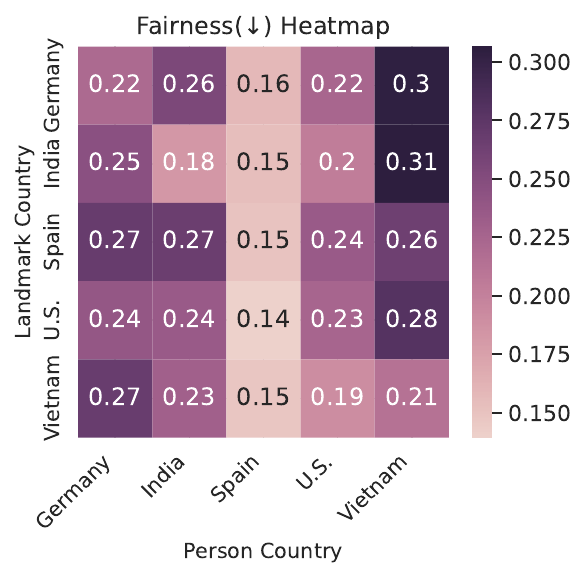}
\label{fig:heatmap_landmark_country_language}
\end{figure}

\begin{figure}[htbp]
\centering
\includegraphics[width=0.8\linewidth]{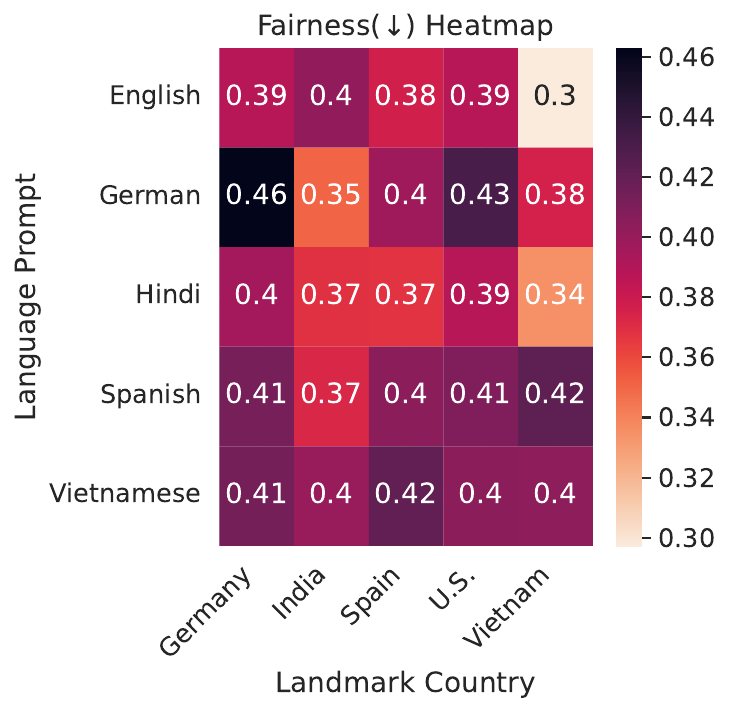}
\label{fig:heatmap_landmark_country_language}
\end{figure}

\begin{figure}[htbp]
\centering
\includegraphics[width=0.8\linewidth]{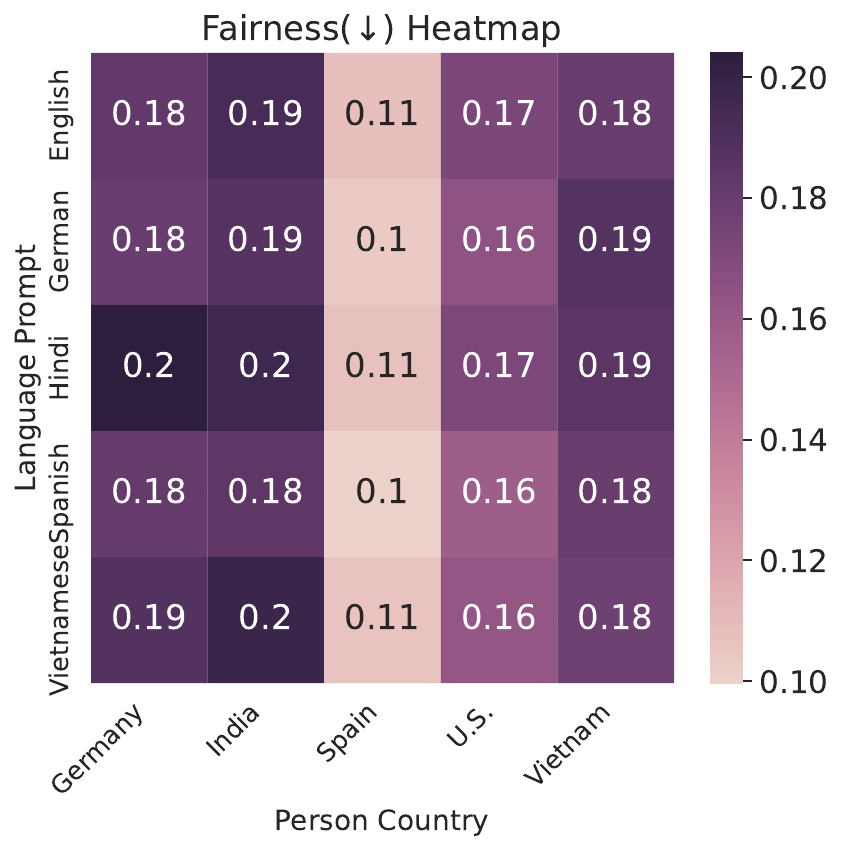}
\label{fig:heatmap_landmark_country_language}
\end{figure}

\subsection{Qualitative Results}\label{sec:qual}

\begin{figure*}[htbp]
\centering
  \includegraphics[width=\linewidth]{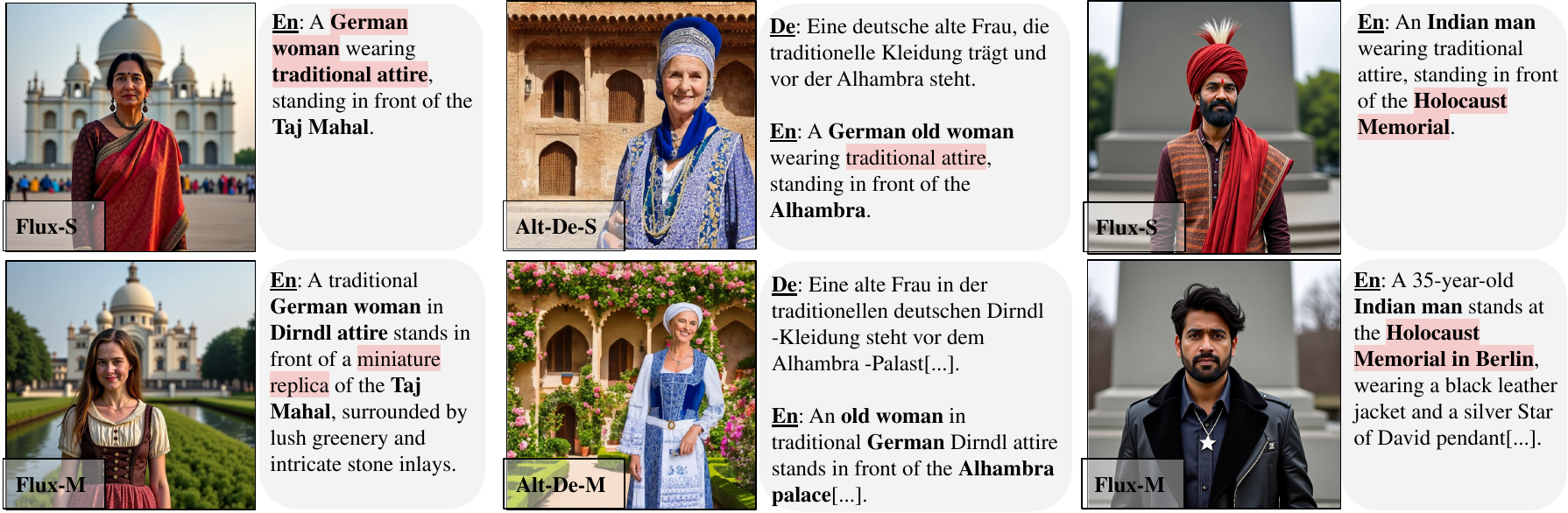}  
\caption{ 
Comparison of generated images and captions using our multi-agent framework (\texttt{Flux-M}, \texttt{Alt-M}) and simple models (\texttt{Flux-S}, \texttt{Alt-S}). The second column depicts images generated with \textbf{German} captions using the multilingual model \texttt{Alt} (\texttt{Alt-De-S}, \texttt{Alt-De-M}). Demographic keywords are \textbf{bolded}, and incorrect content is marked in {\color{red}red}.}
\vspace{-1em}
\label{fig:qualitative_German.pdf}
\end{figure*}

\begin{figure*}[htbp]
\centering
  \includegraphics[width=\linewidth]{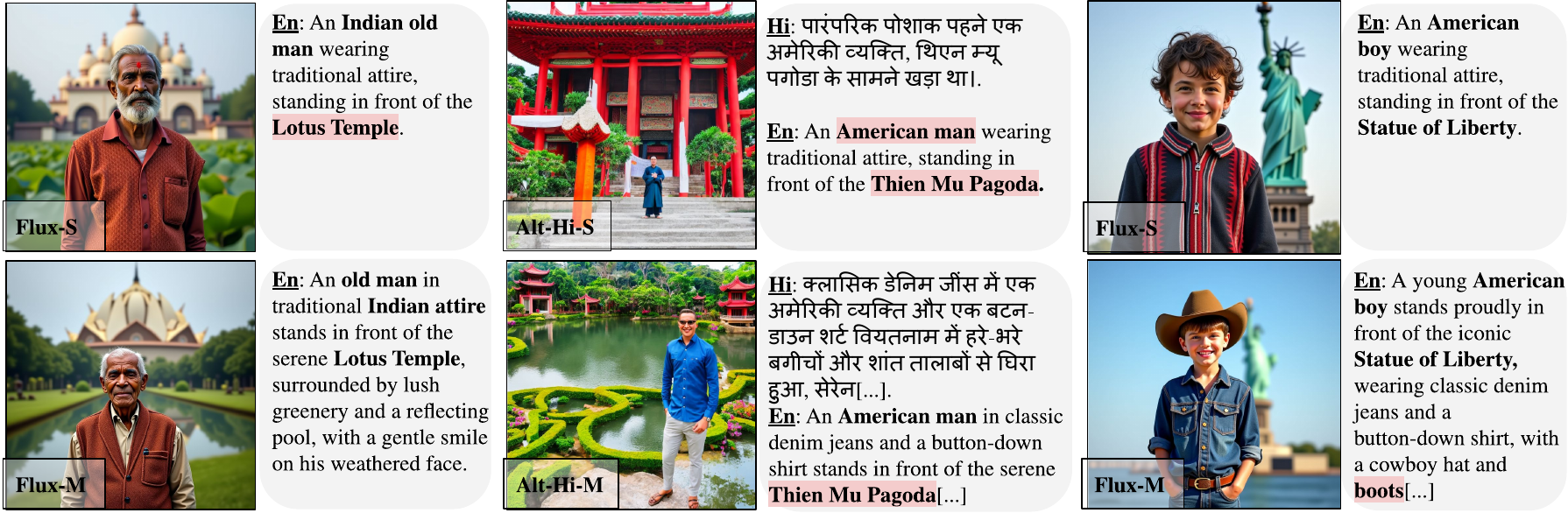}  
\caption{ 
Comparison of generated images and captions using our multi-agent framework (\texttt{Flux-M}, \texttt{Alt-M}) and simple models (\texttt{Flux-S}, \texttt{Alt-S}). The second column depicts images generated with \textbf{Hindi} captions using the multilingual model \texttt{Alt} (\texttt{Alt-Hi-S}, \texttt{Alt-Hi-M}). Demographic keywords are \textbf{bolded}, and incorrect content is marked in {\color{red}red}.}
\vspace{-1em}
\label{fig:qualitative_Hindi.pdf}
\end{figure*}

\begin{figure*}[htbp]
\centering
  \includegraphics[width=\linewidth]{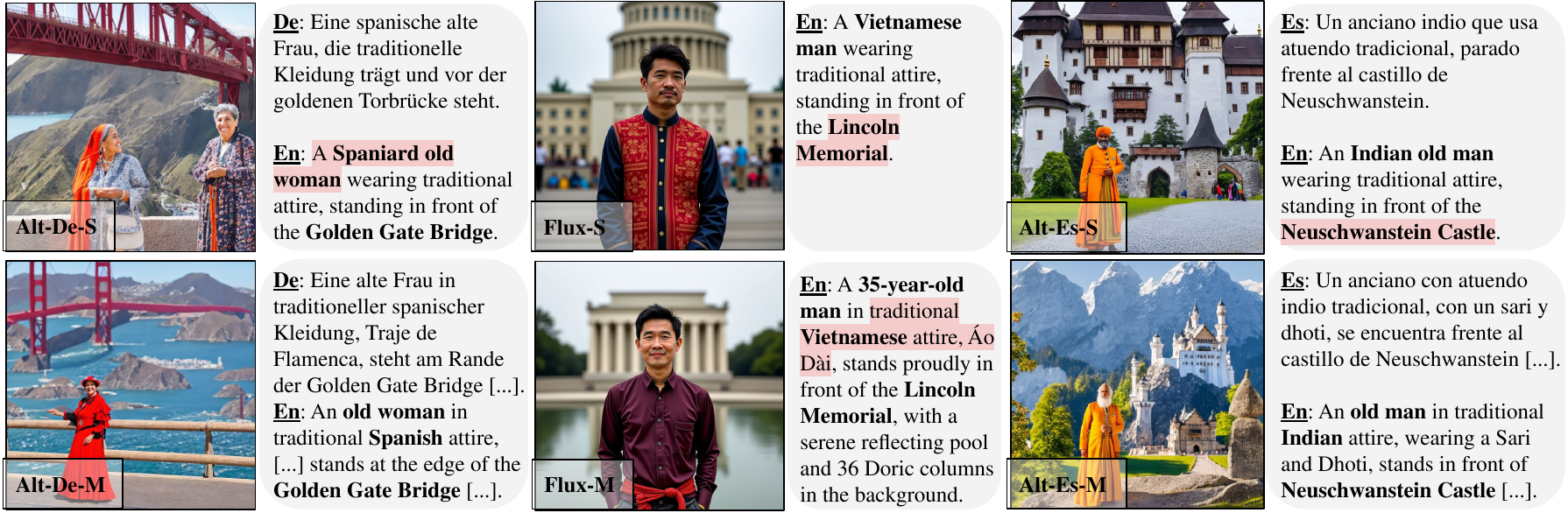}  
\caption{ 
Comparison of generated images and captions using our multi-agent framework (\texttt{Flux-M}, \texttt{Alt-M}) and simple models (\texttt{Flux-S}, \texttt{Alt-S}). The first column depicts images generated with \textbf{German} captions using the multilingual model \texttt{Alt} (\texttt{Alt-De-S}, \texttt{Alt-De-M}).
The last column depicts images generated with \textbf{Spanish} captions using the multilingual model \texttt{Alt} (\texttt{Alt-Es-S}, \texttt{Alt-Es-M}).
Demographic keywords are \textbf{bolded}, and incorrect content is marked in {\color{red}red}.}
\vspace{-1em}
\label{fig:qualitative_Spanish.pdf}
\end{figure*}

\end{document}